\documentclass[11pt]{article}

\usepackage[preprint]{acl}

\usepackage{times}
\usepackage{latexsym}
\usepackage{booktabs}
\usepackage{subcaption}
\usepackage{amsmath}
\usepackage{amssymb}
\usepackage{amsthm} 
\usepackage{cleveref}
\usepackage[T1]{fontenc}

\usepackage[utf8]{inputenc}

\usepackage{microtype}

\usepackage{inconsolata}

\usepackage{graphicx}

\title{Performance, Efficiency and Collapse -- Advantages and Challenges in Offline Post-training of Code LLMs}

\author{
 \textbf{Abhinav Anand\textsuperscript{1, *}} ,
 \textbf{Sanjana Reddy Pachika\textsuperscript{1, *}} ,
 \textbf{Shweta Verma\textsuperscript{1}},
 \textbf{Mira Mezini\textsuperscript{1,2,3}}
 \\
 \textsuperscript{1}TU Darmstadt,
 \textsuperscript{2}Hessian Center for Artificial Intelligence, Darmstadt, Germany,
 \\
 \textsuperscript{3}National Research Center for Applied Cybersecurity ATHENE
 \\
  \textsuperscript{*}Equal Contribution,
\\
 \small{
   \textbf{Correspondence:} \href{mailto:email@domain}{abhinav.anand@tu-darmstadt.de}
 }
}

\begin{document}
\maketitle
\begin{abstract}
Post-training with reinforcement learning (RL) is a critical phase in the development of code-generating large language models (LLMs), as it ensures adherence to instructions and the production of functionally correct code. This process typically requires computationally intensive code sample generation from Transformer-based LLMs and substantial GPU-CPU communication for sequence verification. To address these computational challenges, this work examines whether RL-based post-training can be performed entirely offline by leveraging existing datasets rather than generating new samples. The findings indicate that, with only a few hours of training, zero-shot code generation performance of LLMs can be substantially improved without online sampling. Additionally, offline RL produces performance gains across models ranging from 0.5B to 7B parameters, although the extent of improvement varies among model families.

\end{abstract}

\section{Introduction}
The performance of Transformer-based models on code generation has improved significantly in recent years \cite{cllm_survey}. This improvement is primarily attributed to two factors: (1) the availability of large code datasets derived from open-source repositories \cite{stack, stackv2} and (2) the efficient parallel training capabilities of the Transformer architecture \cite{attention}, which enable the use of extensive datasets during pretraining.

\begin{figure}[htbp]
  \centering
  \includegraphics[width=0.45\textwidth]{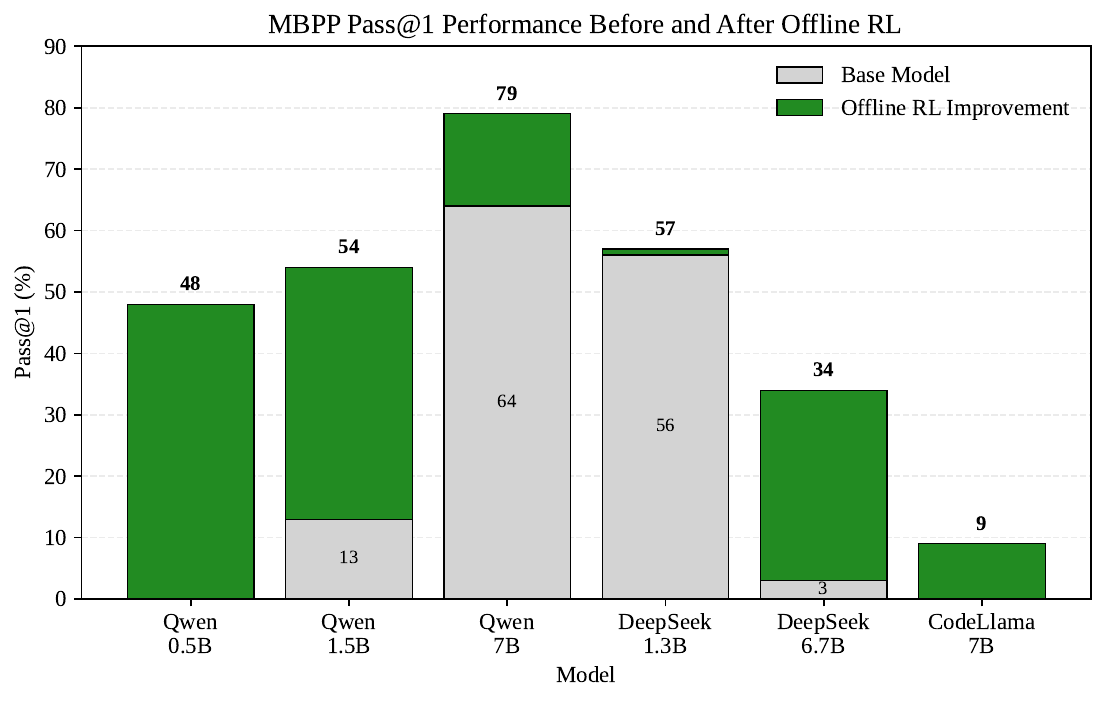}
  \caption{Pass@1 on MBPP benchmark across model families and sizes show that Offline RL with reward can significantly improve the models.}
  \label{fig:mbpp}
\end{figure}

However, pretraining alone is insufficient, and models require post-training with Reinforcement Learning with Verifiable Rewards (RLVR) \cite{rlvr, deepseekr1} to ensure generation of functionally correct code. The post-training process introduces two challenges.
First, pre-existing data is not being used; instead, new data must be sampled from the model or policy under training. Second, the Transformer architecture exhibits slow and inefficient inference. The slow inference combined with GPU-CPU communication for verification of generated sample results in prolonged training times and substantial computational demands. 

Thus, the requirement for online sampling of data from the model constitutes a significant bottleneck, rendering the post-training process computationally expensive and slow. 
As a result, models are typically trained with only one feedback signal, namely functional correctness. Other critical characteristics, such as code efficiency and security, are often neglected because their inclusion would introduce additional computational overhead.

To address the challenges, 
this work investigates whether post-training with verifiable feedback is feasible without online sampling. To this end, we perform post-training with pre-existing dataset and provide an in-depth analysis of offline RL \cite{offlinetut} for code LLMs with experiments across multiple model families, model sizes, learning rate, training epochs and study the stability of offline RL training. 

We study three model families — Qwen Coder \cite{qwencoder} (0.5B, 1.3B, 7B), DeepSeek Coder \cite{deepseekcoder} (1.3B, 6.7B) and CodeLLaMa \cite{codellama} (7B). We utilize the CodeNet data set, which contains problems with multiple correct and incorrect solutions, as well as the execution status of each solution. As shown in Figure \ref{fig:mbpp}, our results demonstrate that training exclusively with offline data can significantly improve the performance of code models across model families and sizes. 

However, model performance is highly sensitive to learning rates and training epochs. While performance improves with multiple training epochs on the CodeNet dataset, prolonged training can result in model collapse.
We analyze the cause of model collapse during RL training. In online RL, instability arises primarily from variance in the advantage function; this variance can be controlled during offline RL. However, using offline data introduces substantial variance in logits, leading to instability after several training epochs. Implementing early stopping based on logit variance and the difference between the predicted logit and the logit assigned to the dataset token can mitigate model collapse.

This work makes the following contributions.
\begin{itemize}
    \item We investigate offline RL across diverse model families and a wide range of model sizes, showing that models with different sizes and architectures derive significant benefits from offline RL.
    \item We examine the impact of learning rate (LR) and training over multiple epochs on offline RL training. We demonstrate that training outcomes are highly sensitive to the choice of LR. The performance can improve with training over multiple epochs. However, prolonged training can also lead to model collapse.
    \item We analyze the causes of instability in offline RL training. Unlike online RL, where instability primarily arises from advantage variance, we find that logit variance is the main source of instability in offline RL. We further investigate the factors contributing to the increase in logit variance during training and propose diagnostics for early stopping.
    
\end{itemize}

To the best of our knowledge, this is the first work to conduct an in-depth analysis of offline RL for code LLMs. Our findings suggest a pathway toward more compute- and data-efficient post-training of code LLMs. All training and evaluation reported in this paper are performed using a single GPU to ensure efficient offline training -- additional improvements may be realized by employing larger batch sizes with multiple GPUs. Additionally, we perform full model training. 
Although our motivation encompasses multiple types of feedback, including efficiency and security, beyond correctness, the present analysis focuses on offline RL with functional correctness as the feedback signal. The extension to multiple feedback types is identified as a direction for future research.

\section{RL Framework}
Reinforcement Learning (RL) is commonly used to align LLMs with specific desirable characteristics. For LLMs that generate code, this characteristic typically involves adherence to the intended functionality, which is assessed using test cases. 

\subsection{RL Algorithm}
Several RL algorithms, including RLOO \cite{rloo, rloobuy}, GRPO \cite{grpo}, and PPO \cite{ppo}, have been proposed in the literature. In this work, our objective is to validate RL for LLMs in an offline setting, not to achieve state-of-the-art results; therefore, we use the simplest algorithm: RLOO (see Appendix \ref{RLOO} for details about RLOO). Employing a simple algorithm reduces the need for wide range of RL-specific hyperparameters and provides a conservative estimate of LLM performance. Alternative algorithms may further enhance performance.


\subsection{Offline Reinforcement Learning} \label{sec:rl}
In online RL, the agent continuously interacts with the environment to generate new training samples. In contrast, offline RL \cite{offlinetut} utilizes a fixed dataset of previously collected experiences. This dataset typically includes prompts, generated programs, and their associated rewards, enabling policy optimization without further interaction with the environment.

\begin{figure}[t]
\centering
\includegraphics[width=\columnwidth]{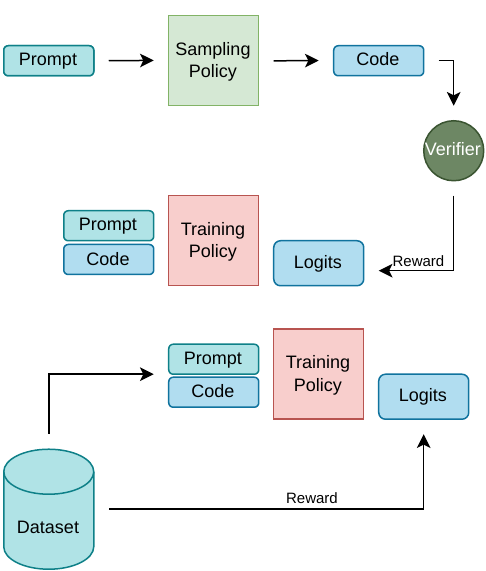}
\caption{Comparison between Online (top) and Offline (bottom) RL of LLMs.}
\label{fig:offline v online}
\end{figure}

In the context of code generation, online RL typically requires the model to generate candidate programs, execute them against a test suite, compute rewards, and update the policy iteratively. Although this approach has been successfully applied to enhance large language models, it remains computationally expensive and slow because of the repeated program generation and execution during training, as well as the inherent inefficiency of the transformer architecture.

Offline RL mitigates these computational costs by learning directly from pre-existing datasets of code submissions and their associated execution outcomes. Because rewards are already available, training can proceed without repeatedly generating and evaluating new programs (See Figure \ref{fig:offline v online} for the contrast between online and offline RL).

\begin{figure}[t]
\centering
\includegraphics[width=\columnwidth]{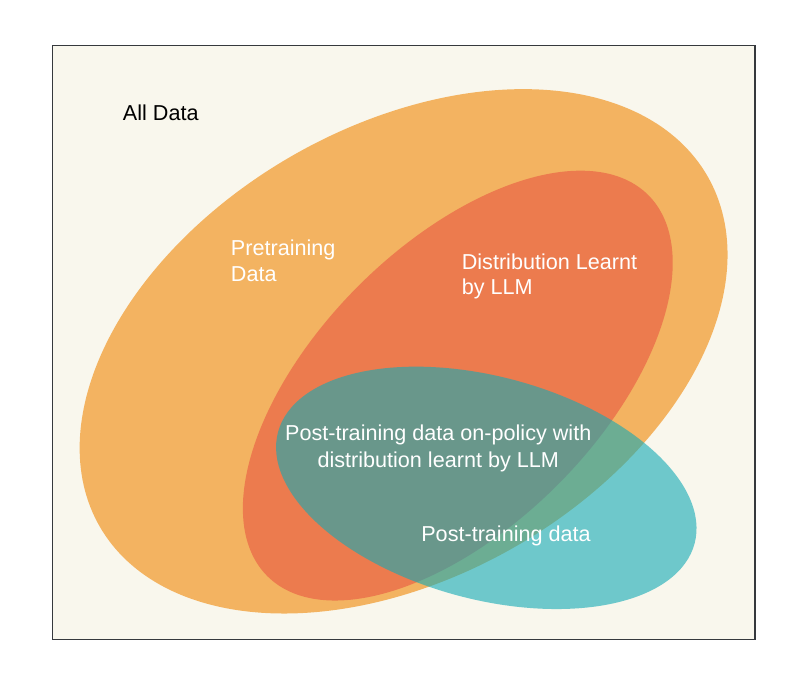}
\caption{LLMs are pre-trained with human data and approximate a policy that generates human data. Due to the pre-training on web scale data, it is possible that offline human written code is on-policy with respect to the LLM.}
\label{fig:why offline}
\end{figure}

In this work, we adopt the offline RL setting by optimizing the policy using a fixed dataset, CodeNet \cite{codenet}, which consists of code submissions and rewards derived from their recorded execution outcomes. The resulting policy is subsequently evaluated on unseen code generation benchmarks to assess its ability to generalize beyond the training data.

For each training group, the RLOO leave-one-out baseline is computed by averaging the rewards of the remaining samples,


\begin{equation}
b_i=\frac{1}{K-1}\sum_{j\neq i} r_j,
\end{equation}

where $K$ is the size of the group and $r_i$ denotes the reward assigned to the $i$-th sample. The corresponding advantage is then computed as

\begin{equation}
A_i=r_i-b_i.
\end{equation}

Rather than directly using the raw advantages for policy optimization, we further apply group-wise advantage normalization following the normalization strategy employed in GRPO,

\begin{equation}
\hat{A}_i=\frac{A_i-\mu_A}{\sigma_A+\epsilon},
\end{equation}

where $\mu_A$ and $\sigma_A$ are the mean and standard deviation of the advantages within each training group, and $\epsilon$ is a small constant added for numerical stability. This normalization reduces the effect of varying the reward scales between different groups and leads to a more stable optimization.

Normalized advantages are then used to optimize the policy through the REINFORCE objective \cite{reinforce},

\begin{equation}
L=-\frac{1}{N}\sum_{i=1}^{N}\hat{A}_i\log\pi_\theta(y_i\mid x_i),
\end{equation}

where $N$ is the number of samples in the batch. The resulting objective combines the leave-one-out advantage estimation of RLOO with GRPO-style advantage normalization, providing a stable optimization procedure for our offline RL framework.

\subsection{Why Policy Gradient?} \label{hypothesis}

Traditionally, offline RL is performed using Q-learning \cite{offlinetut}. However, we employ a policy gradient algorithm, originally developed for online policy settings, in the offline setting. Our reasoning for doing so is visually explained in Figure \ref{fig:why offline} . Since LLMs are already pre-trained on human-generated data and not trained with RL from scratch, we contend that the LLM policy remains close to existing human-generated reward datasets, such as CodeNet. Given that both the pre-training and reward datasets are human-generated, and the policy is initialized with the pre-training dataset, we posit that offline training is effectively on-policy. For samples where this assumption is valid, the model can learn from feedback in the offline setting.

Additionally, \citet{ILQL} used Q-learning to train a value head on offline data and rewards. However, the authors observed that extracting the policy from the value head during inference results in instability for LLMs. Since the policy in Q-learning is implicit, it is difficult to explore the sources of instability. Consequently, we prioritize learning the policy directly during training.

\section{Study Methodology}

\subsection{Models}
We evaluated our approach on six decoder-only code generation models from three model families: Qwen2.5-Coder \cite{qwencoder}, DeepSeek-Coder \cite{deepseekcoder}, and CodeLlama \cite{codellama}. The selected models range from 0.5B to 7B parameters and are widely used open-source model families. For all experiments, we used the publicly available base models and applied offline RL directly without supervised fine-tuning or instruction tuning.

\subsection{Dataset}
Our offline RL dataset is constructed from the CodeNet dataset \cite{codenet}. To align with the evaluation benchmarks used in this work, the training data consists exclusively of Python submissions. To mitigate the imbalance caused by problems with a large number of submissions, we retain a maximum of 20 submissions per problem.

The final dataset comprises 8,321 training samples from 609 unique programming problems. Each sample includes a programming problem, a Python submission, and its recorded execution status. Accepted and incorrect submissions are incorporated, allowing the model to learn from both successful and unsuccessful programming attempts. During training, execution outcomes are assigned to scalar rewards, which are utilized by the offline RL algorithm.
\begin{figure*}[h]
\centering
    \begin{subfigure}[b]{0.99\textwidth}
     \includegraphics[width=0.33\linewidth]{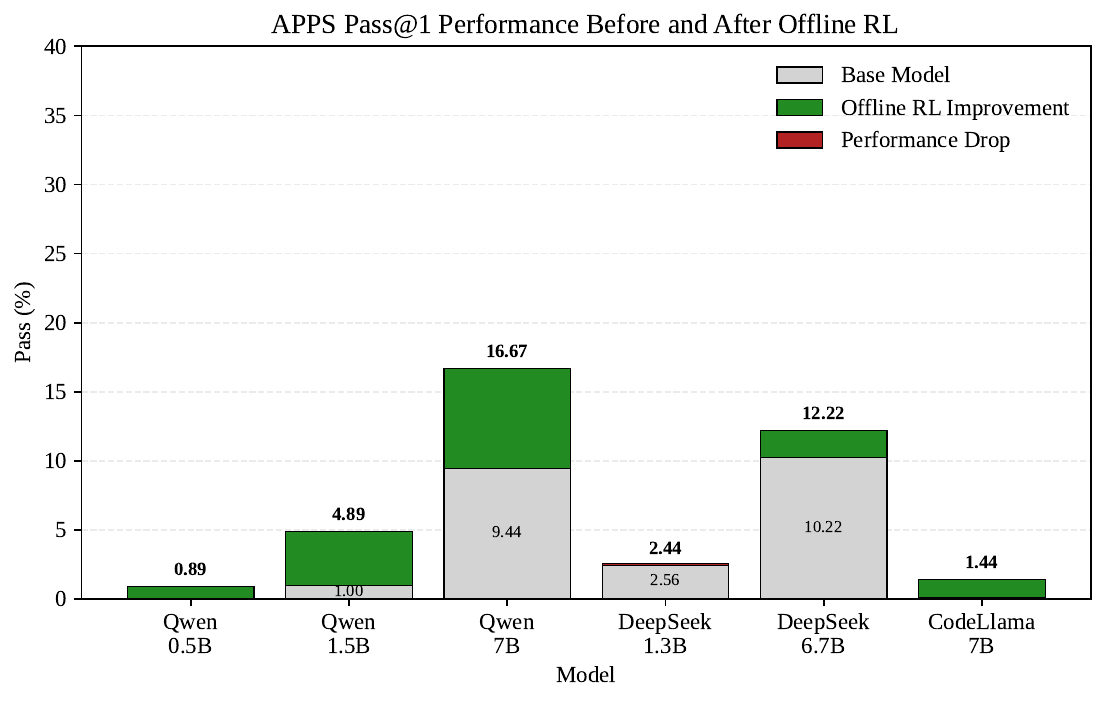} 
     \includegraphics[width=0.33\linewidth]{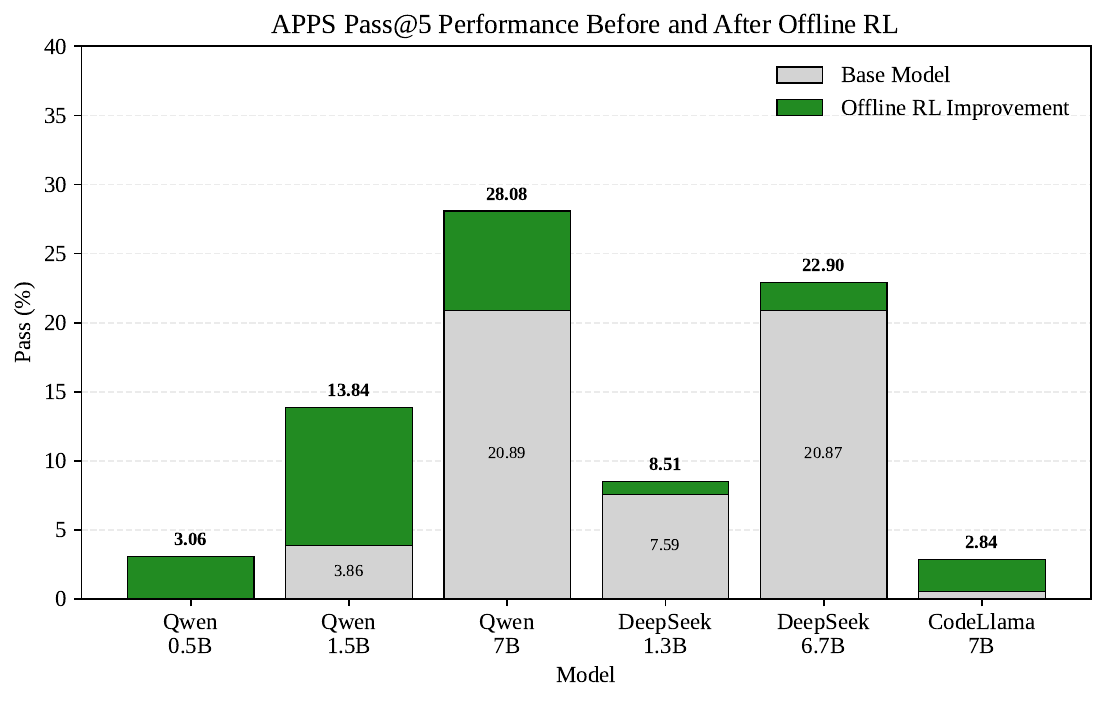} 
     \includegraphics[width=0.33\linewidth]{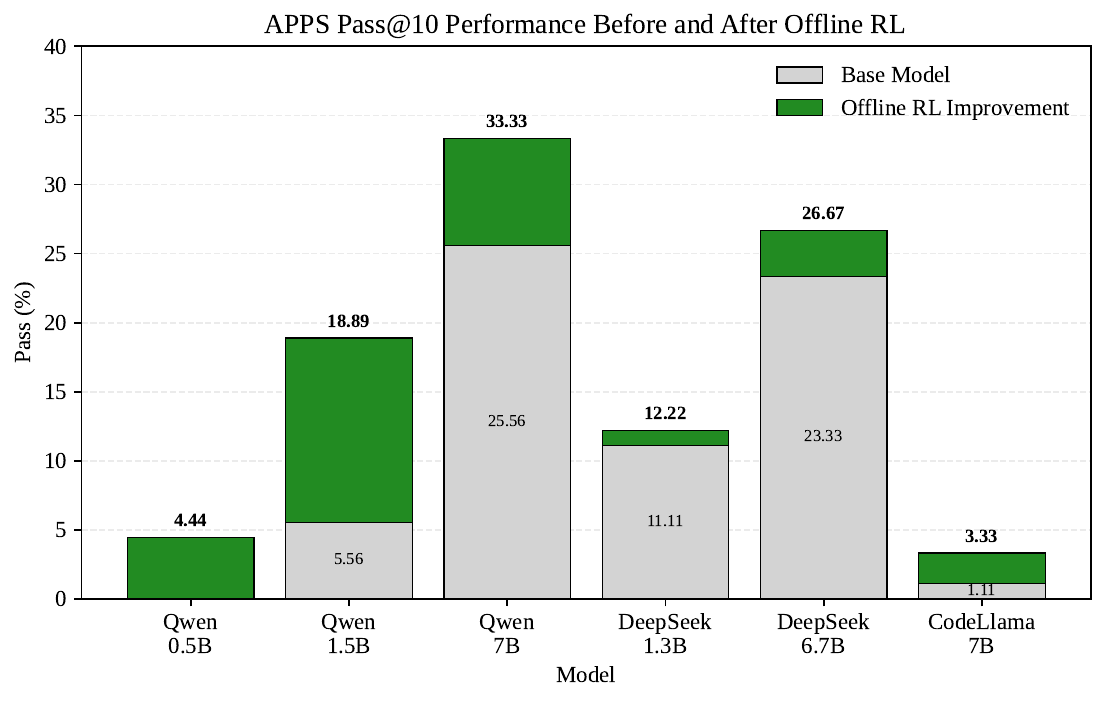}
    \end{subfigure}
\caption{Performance in terms of pass@1, pass@5 and pass@10 on APPS dataset.}
\label{fig:apps}
\end{figure*}

\subsection{Training Setup}
All models were trained for 10 epochs using the offline RL framework described in Section \ref{sec:rl}. Training consistently used a group size of four and eight gradient accumulation steps. The models were trained with different learning rates to identify optimal optimization settings for each model.

The reward, r, for a code status is given as
\begin{equation}\label{eq:reward_function}
r =
\begin{cases}
+1.0 & \text{All Test Cases Passed} \\
-0.1 & \text{Test Cases Failed} \\
-0.5 & \text{Time Limit Exceeded} \\
-0.5 & \text{Runtime Error} \\
-1.0 & \text{Compile Error} 
\end{cases}
\end{equation}

The variance in advantage represents a significant factor that contributes to instability in RL training. Given that all samples corresponding to a specific prompt are available, groups are constructed to minimize variance. We ensure that each group contains one correct sample (r = 1) and one incorrect sample (r = -1), while the remaining two samples may have any execution status. This grouping strategy reduces the advantage variance.

\subsection{Computational Constraints}
All experiments were conducted on a single 80GB A100 GPU. Since the evaluated models range from 0.5B to 7B parameters, the batch size was adjusted based on the model size to accommodate GPU memory constraints. The duration of training depends on the specific size of the model.

\subsection{Benchmarks and Metric}
We evaluate the model's ability to generate Python code on two benchmarks -- MBPP \cite{mbpp} and APPS \cite{apps}. MBPP comprises simple Python problems; therefore, only pass@1 is reported for this benchmark. APPS includes problems of varying difficulty levels (introductory, interview, and competitive). We report pass@1, pass@5, and pass@10 for each difficulty level.

\section{Results and Discussion}
In this section, we present the results of the model evaluation. 

\subsection{Performance}
Figure \ref{fig:mbpp} presents the MBPP Pass@1 results for the evaluated models. Offline RL substantially enhances model performance across different model sizes. Specifically, Qwen 0.5B shows a 48\% increase, while Qwen 7B and DeepSeek 6.7B improve by 15\% and 31\%, respectively. Although CodeLlama also shows improved performance, its overall result remains low at 9\%. This poor performance may be attributed to a misalignment between the pretraining data distribution learned by the model and the post-training dataset.


Figure \ref{fig:apps} shows that performance improves in the more challenging APPS dataset with pass@1, pass@5 and pass@10. Furthermore, we observe that the improvements occur consistently at all three difficulty levels (see Figure \ref{fig:apps_difficulty} in Appendix). The smallest model, Qwen 0.5B, also exhibits improved performance in competition-level problems for pass@1, pass@5, and pass@10. Table \ref{tab:comp} presents a comparison of performance improvements in very small models using CodeRL \cite{coderl}, PPOCoder \cite{ppocoder}, and our proposed Qwen 0.5B training. Offline training produces greater improvements in small models, and the difference being particularly pronounced in competition-level problems.

\begin{table}[t]
\centering
\tiny
\begin{tabular}{lc|ccc|ccc}
\hline
\textbf{Model} & \textbf{size}& \multicolumn{3}{c|}{\textbf{pass@1}} & \multicolumn{3}{c}{\textbf{pass@5}} \\
\hline
CodeRL & 770M & 1.3 & 0.16 & 0.3 & 4.30 & 1.27 & 0.8 \\
PPOCoder & 770M & 1.6 & 0.1 & 0.3  & 4.8 & 1.13 & 1.0 \\ 
Qwen (ours) & 490M & \textbf{1.7} & \textbf{0.3} & \textbf{0.7} & \textbf{4.9} & \textbf{1.7} & \textbf{2.6}\\
\hline
\end{tabular}
\caption{Improvement over base model. CodeRL and PPOCoder were trained with APPS dataset while we train on CodeNet and do zero-shot evaluation on APPS.}
\label{tab:comp}
\end{table}

CodeLlama demonstrates substantial improvement in introductory-level problems; however, its performance does not improve in interview- or competition-level problems.

We hypothesize that the observed disparity in improvements across model families suggests that the effectiveness of offline RL depends on the composition of the pre-training dataset. Employing more effective heuristics for sample selection or incorporating samples generated by the model may enhance performance when there is misalignment between pre-training and post-training data.

The learning rate is another factor that influences model performance. The optimal learning rate varies according to the size and family of the model. Generally, lower learning rates do not produce improvement, whereas very high learning rates may initially enhance performance but subsequently cause rapid degradation. Intermediate learning rates tend to produce performance gains over multiple epochs. Figure \ref{fig:lr with epoch} presents the pass@1 in MBPP for various learning rates at different epochs.

\begin{table}[t]
\centering
\tiny
\begin{tabular}{lccc}
\hline
\textbf{Model} & \textbf{Batch Size} & \textbf{Training Time / epoch}& \textbf{Total Training Time} \\
               &                     & \textbf{(minutes)}            &\\
\hline
Qwen 0.5B      &  6 & 3:31 & 35m \\
Qwen 1.5B      &  6 & 15:12 & 2h 32m\\
Qwen 7B        &  1 & 17:23 & 2h 53m  \\
DeepSeek 1.3B  &  6 & 16:30 & 2h 45m\\
DeepSeek 6.7B  &  2 & 33:16 & 5h 32m\\
CodeLlama 7B   &  2 & 38:21 & 6h 23m  \\
\hline
\end{tabular}
\caption{Training time for each model.}
\label{tab:time}
\end{table}

\subsection{Efficiency}
Offline RL eliminates the need for computationally intensive sampling from transformer models and the associated GPU-CPU communication required to run a verifier over generated samples. Table \ref{tab:time} shows that performance improvements are achievable with minimal training on a single GPU.

\subsection{Collapse}
\begin{figure*}[h!]
\centering
    \begin{subfigure}[b]{0.99\textwidth}
     \includegraphics[width=0.33\linewidth]{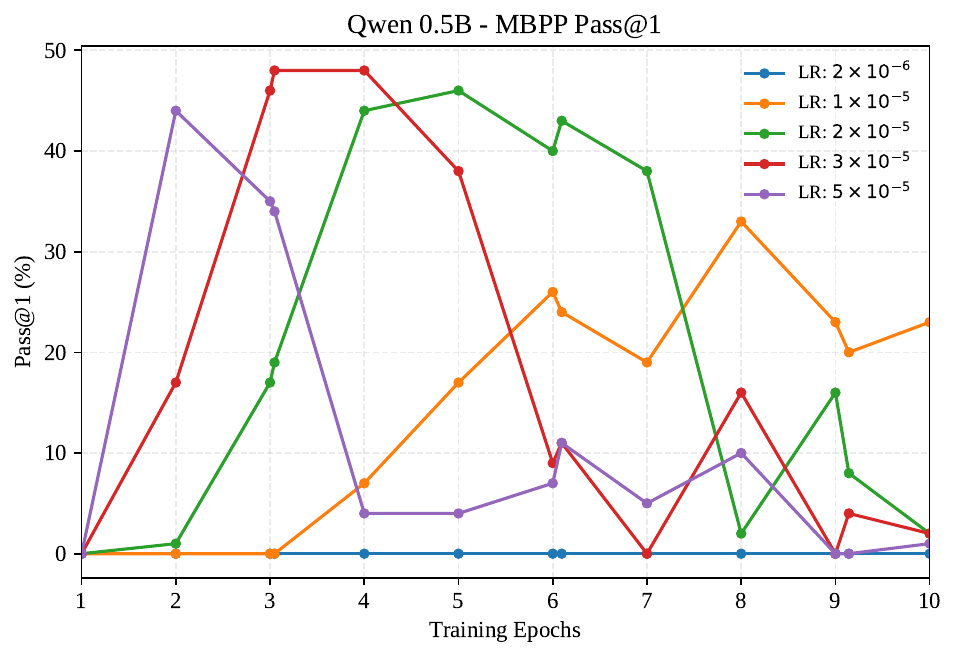} 
     \includegraphics[width=0.33\linewidth]{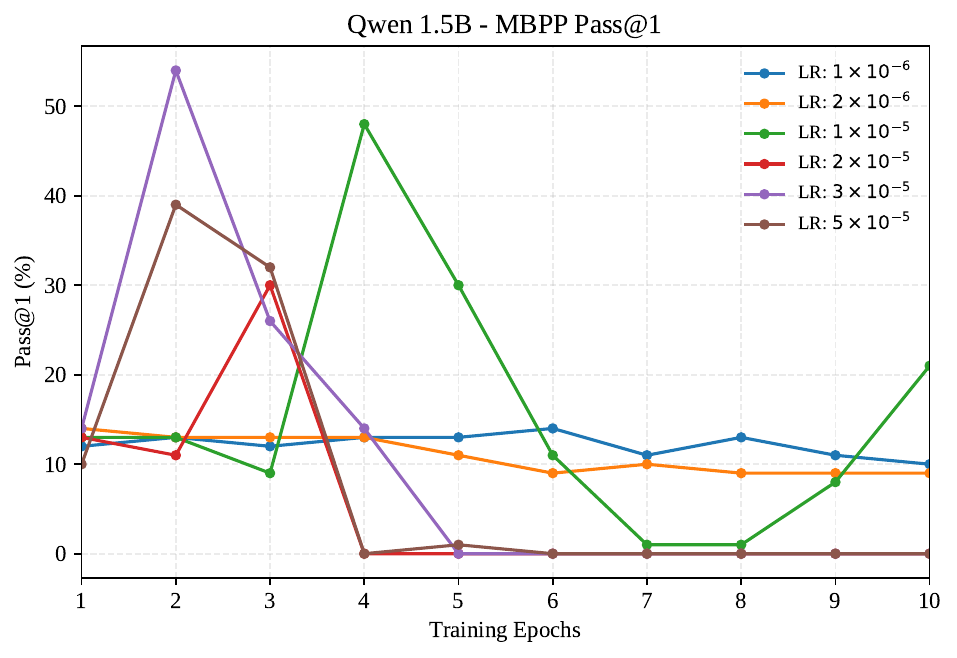}
     \includegraphics[width=0.33\linewidth]{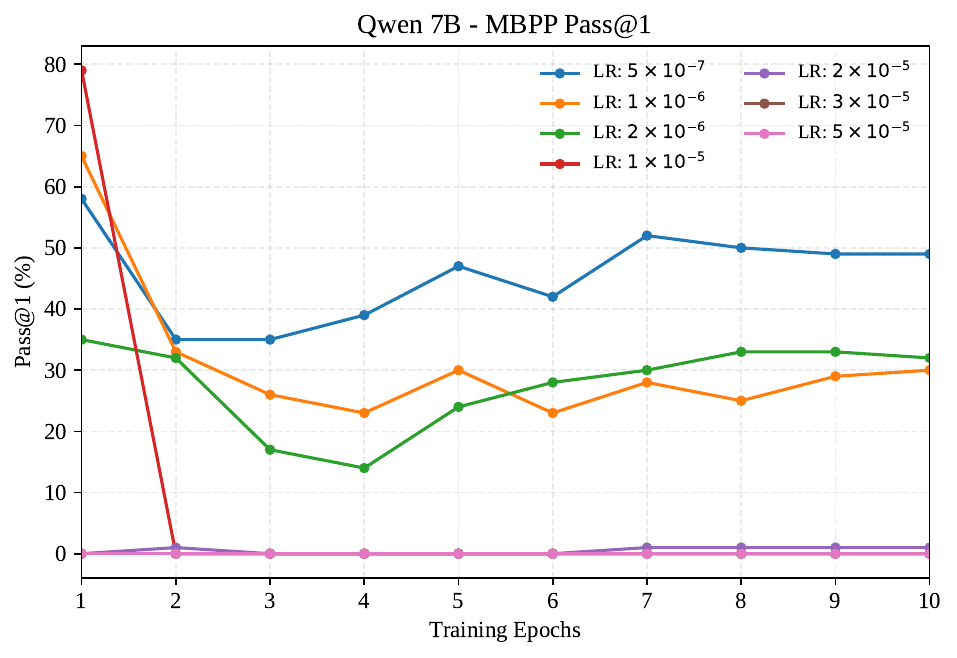}
    \end{subfigure}
    
    \begin{subfigure}[b]{0.99\textwidth}
     \includegraphics[width=0.33\linewidth]{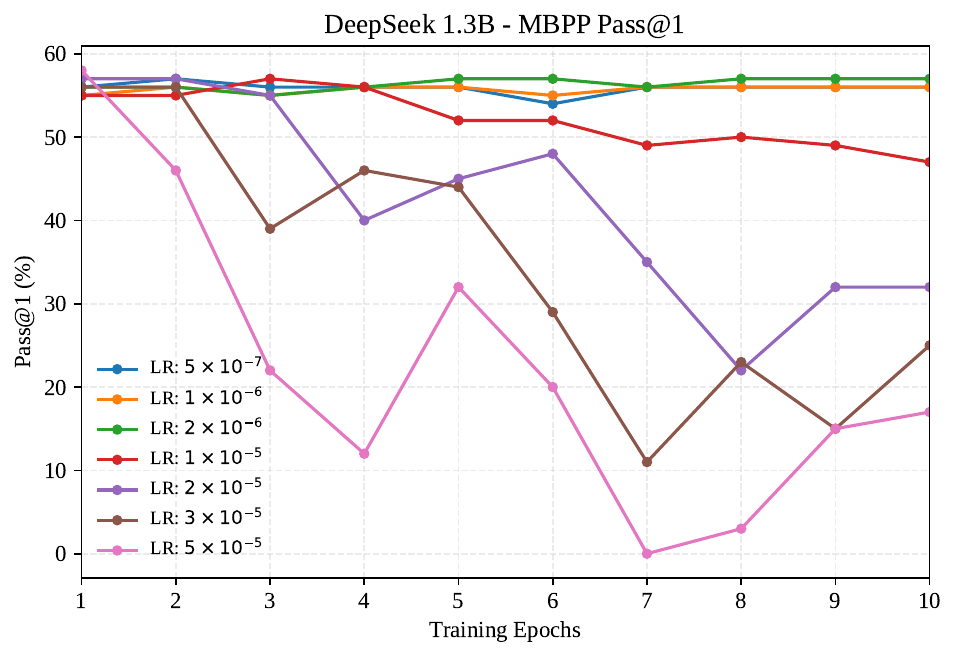} 
     \includegraphics[width=0.33\linewidth]{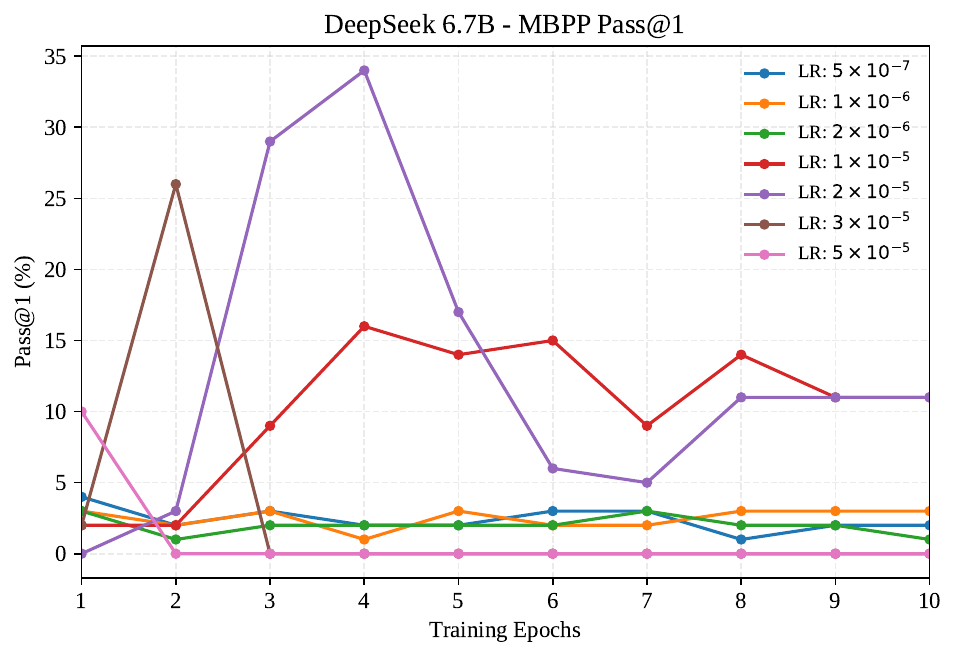}
     \includegraphics[width=0.33\linewidth]{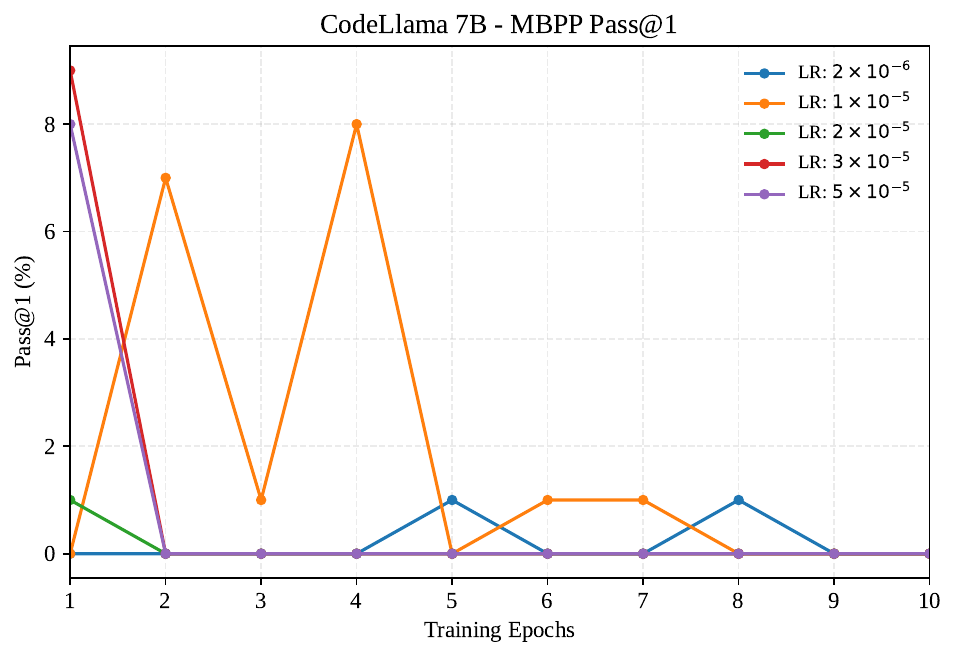} 
    \end{subfigure}
    
\caption{Pass@1 on MBPP with different learning rate over 10 epochs. While the performance improve with some learning rate, prolonged training can result in model collapse or very poor performance.}
\label{fig:lr with epoch}
\end{figure*}

Models require training over multiple epochs to achieve optimal performance. However, extended training often results in model collapse. For example, Qwen 0.5B attains peak performance with a learning rate of $3\times10^{-5}$ after 4 epochs. Further training reduces performance, with the model collapsing to a pass@1 of 0 by the seventh epoch (Figure \ref{fig:lr with epoch}). A similar performance collapse is observed across all model families and sizes.

Performance trends across learning rates and epochs in Figure \ref{fig:lr with epoch} indicate effective post-training strategies with offline data — train using learning rates between $1\times10^{-5}$ and $5\times10^{-5}$, conduct training over multiple epochs and terminate training prior to model collapse.

Different models reach peak performance and experience collapse at varying epochs. To identify the sources of instability that lead to model collapse and to determine optimal stopping points, we analyze the training dynamics.




\section{Instability Analysis}
The RLOO training objective depends on two factors -- advantage and logits. We study the variance in both and report our findings in this section.

\subsection{Advantage Variance}


Figure~\ref{fig:raw_adv_variance} presents the evolution of the variance of the advantages during training. In all training configurations, the variance of the advantage remained stable and did not show a correlation with the progress of training or the performance of the model. Therefore, the advantage remains approximately constant at a fixed value.

Advantage variance is a primary source of instability in online RL. However, our group design, which consistently includes at least one correct sample and one incorrect sample, enables effective control of the variance of the advantage.



\subsection{Log Probability Variance Analysis}

\begin{figure*}[h!]
\centering
    \begin{subfigure}[b]{0.99\textwidth}
     \includegraphics[width=0.33\linewidth]{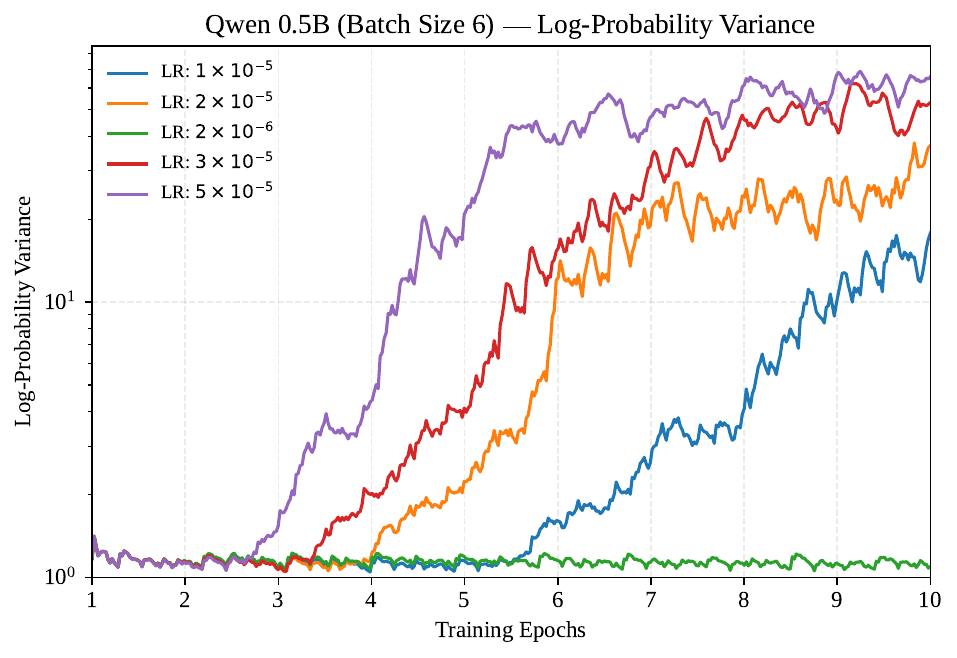} 
     \includegraphics[width=0.33\linewidth]{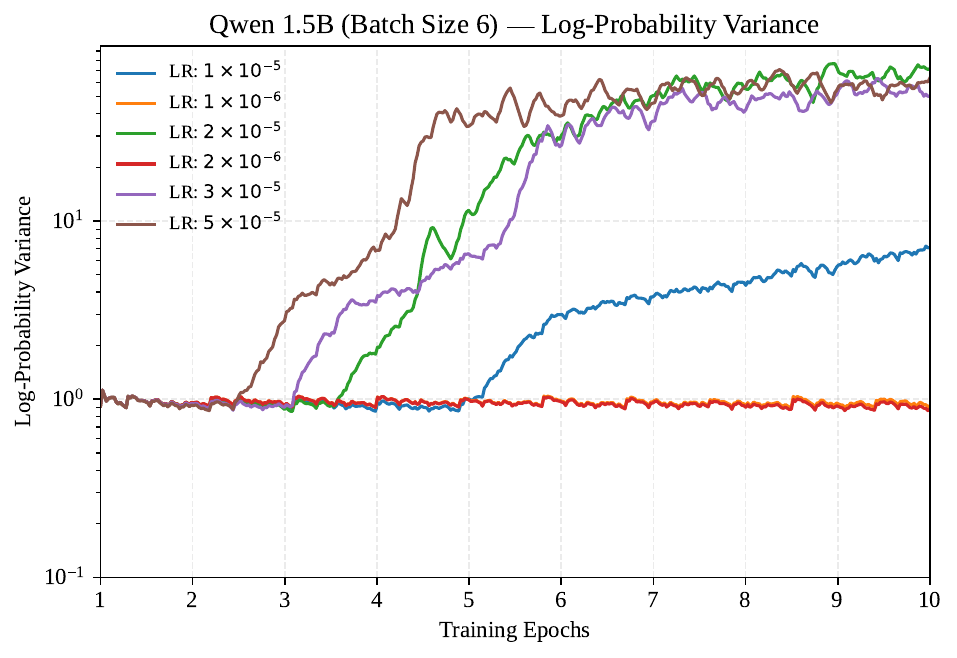}
     \includegraphics[width=0.33\linewidth]{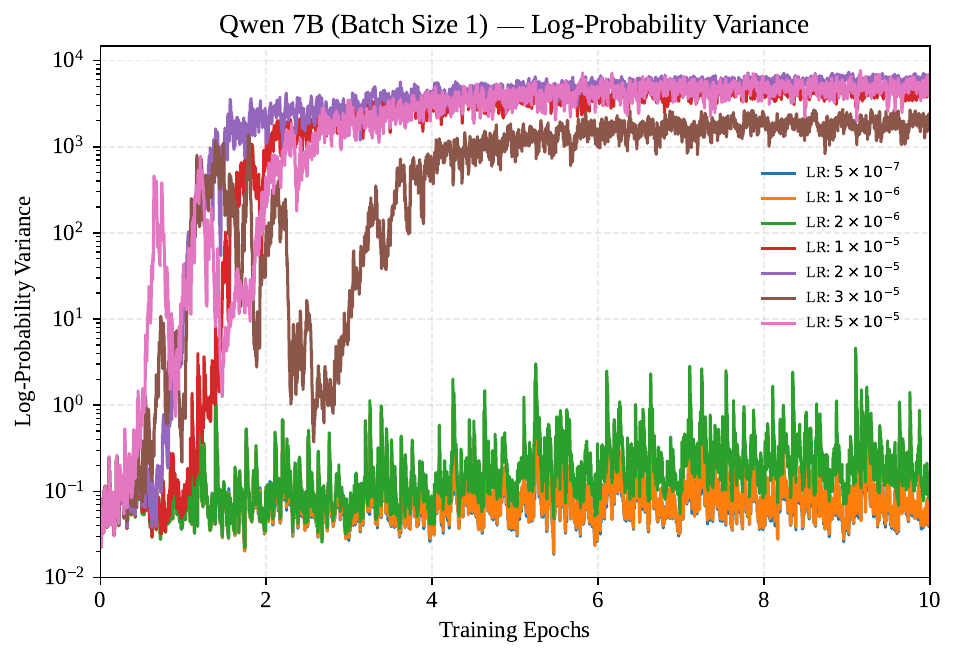}
    \end{subfigure}
    
    \begin{subfigure}[b]{0.99\textwidth}
     \includegraphics[width=0.33\linewidth]{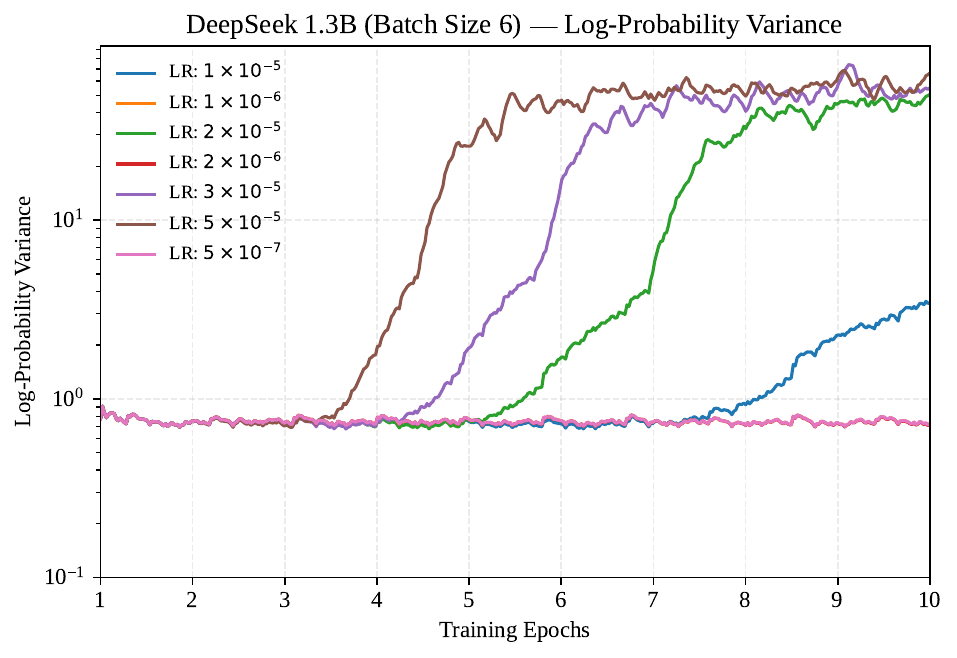} 
     \includegraphics[width=0.33\linewidth]{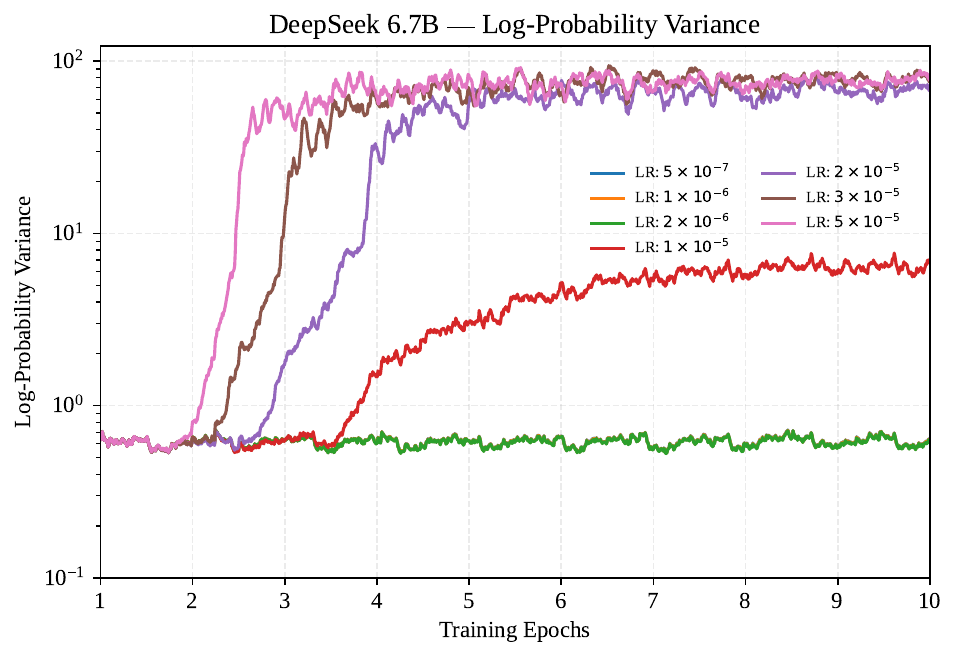}
     \includegraphics[width=0.33\linewidth]{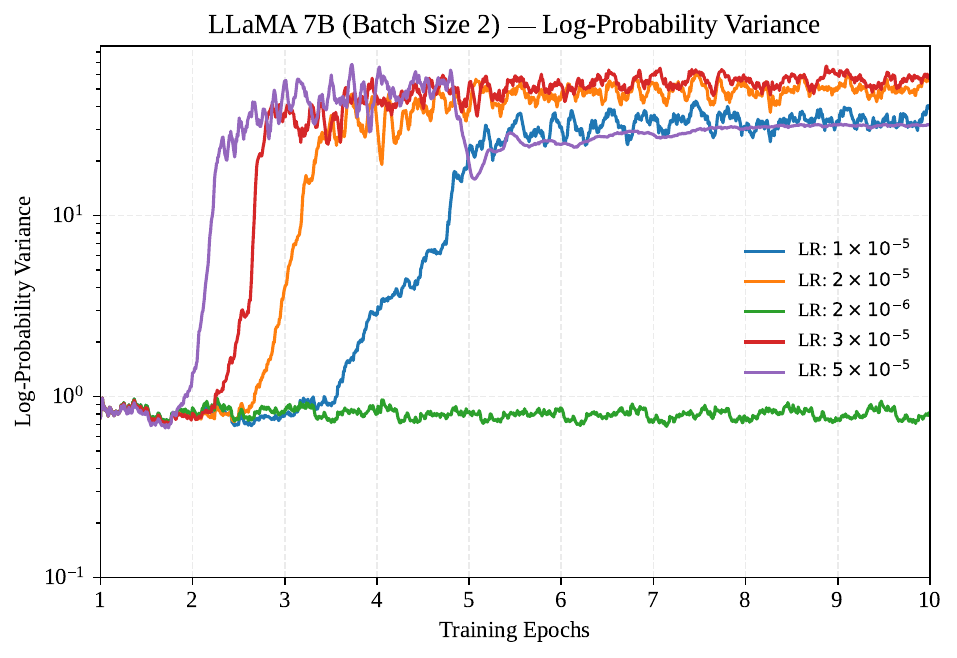} 
    \end{subfigure}

\caption{Variance in logits across training epochs for all models. There are three different variance characteristics -- no variance, early rise and late rise. The characteristic is correlated to performance improvements.}
\label{fig:logprob_variance}
\end{figure*}

\begin{figure*}[h!]
\centering
    \begin{subfigure}[b]{0.99\textwidth}
     \includegraphics[width=0.33\linewidth]{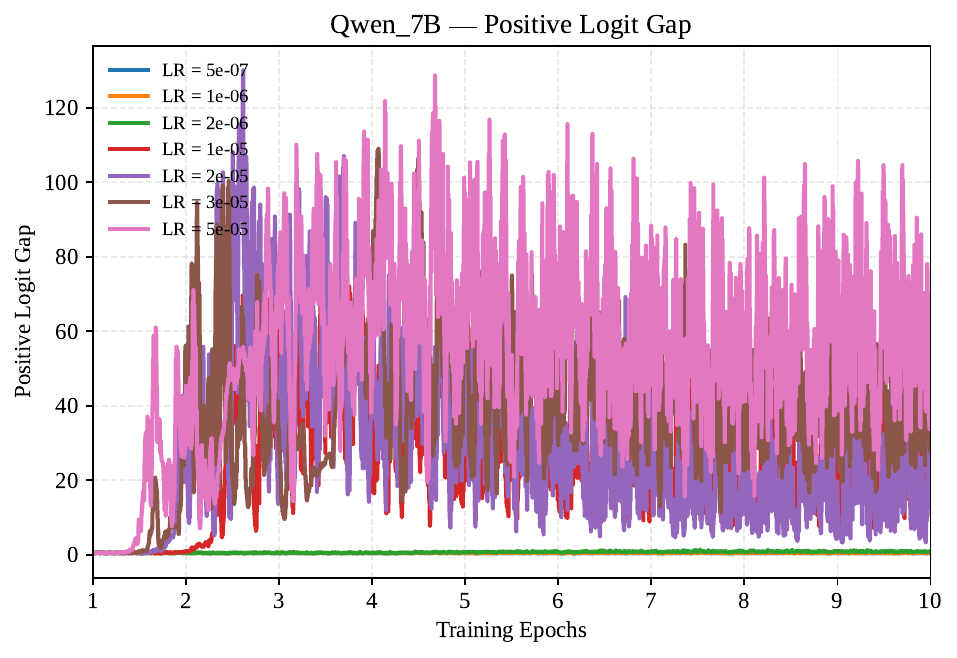} 
     \includegraphics[width=0.33\linewidth]{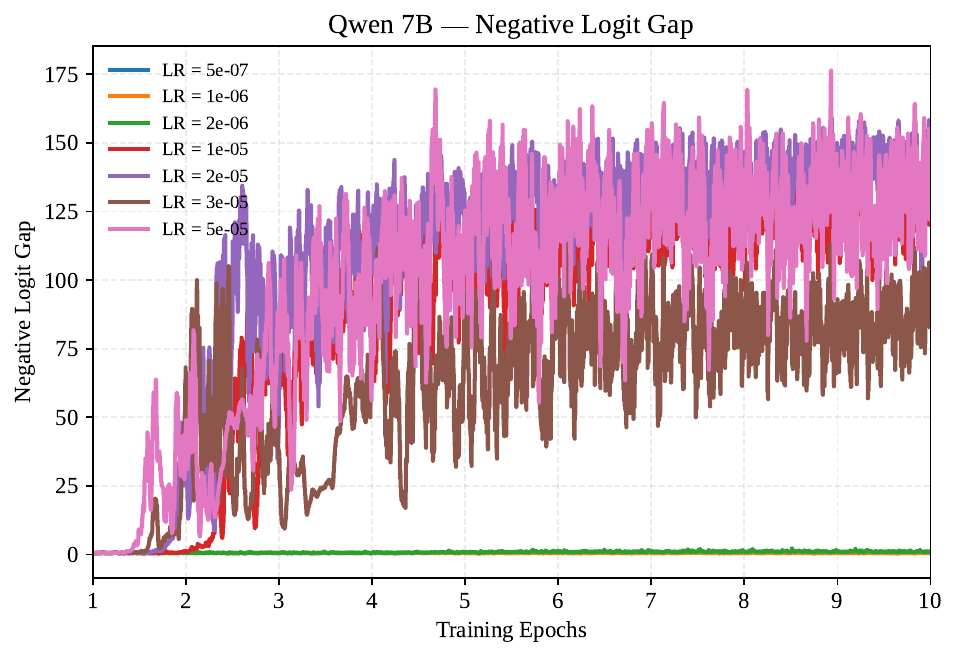}
     \includegraphics[width=0.33\linewidth]{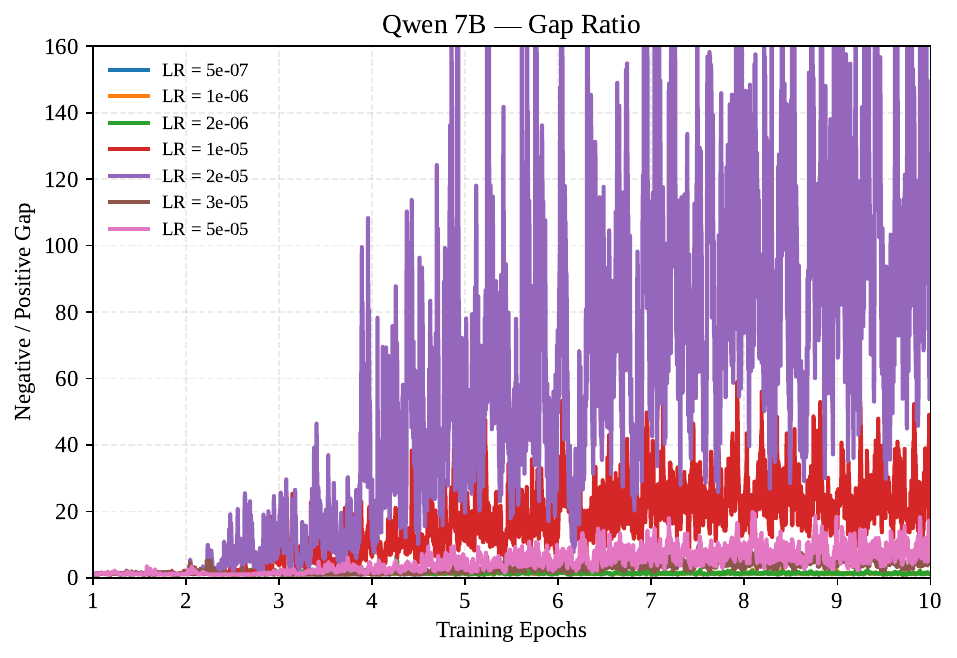}
    \end{subfigure}
\caption{Negative gap, positive gap and gap ration during training of Qwen 7B model.}
\label{fig:logit_gap_qwen}
\end{figure*}

Figure~\ref{fig:logprob_variance} presents the variance in log-probability of tokens from offline data. Lower learning rates result in low and stable log-probability variance throughout training across all evaluated model families. For instance, configurations with learning rates of $1\times10^{-6}$ or lower consistently yield final log-probability variance values below 0.12. In contrast, higher learning rates substantially increase log-probability variance, often by several orders of magnitude. Runs that collapsed to 0\% Pass@1, such as Qwen 1.5B at $2\times10^{-5}$ and DeepSeek 6.7B at $3\times10^{-5}$, exhibited final log-probability variances of 3223.76 and 4094.6, respectively. DeepSeek 1.3B represents an exception; while the base model performs well, offline RL maintains similar performance when variance is low but results in collapse when variance is high (for example, $lr = 3\times10^{-5}$).

Although a large log-probability variance indicates training instability, the relationship between variance and performance is complex. Initially, low variance is associated with model improvement, but as variance increases, it eventually leads to performance collapse in later epochs. In contrast, when the same model is trained with a learning rate of $2\times10^{-6}$, the variance remains low throughout, but the performance of the model does not improve during training.

A similar trend is observed across models: consistently low variance during training or very high initial variance results in poor performance. In contrast, models with an initial low variance that subsequently increases (for example, learning rates of $3\times10^{-5}$ for Qwen 0.5B and 1.5B, $1\times10^{-5}$ for Qwen 7B, $2\times10^{-5}$ for DeepSeek 6.7B, and $3\times10^{-5}$ for CodeLlama) achieve optimal performance before high variance leads to collapse. The only exception is DeepSeek 1.3B, which already shows strong performance on the MBPP task.

Analysis of log-probability variance yields two key observations. First, in contrast to online RL, where instability arises from advantage variance, instability in offline RL originates from variance in logits. Therefore, developing methods to control this variance is essential for improving offline RL performance. Second, increased model performance is accompanied by rising variance, making variance monitoring and early stopping critical to prevent performance collapse.


\subsection{Logit Gap}
A primary challenge associated with using offline data is staleness, which refers to the discrepancy between the offline data and the model's current policy. To evaluate the impact of staleness, this work examines the gap between the preferred token of the model and its prediction of tokens from the offline dataset, as illustrated in Figure \ref{fig:logit gap}.

Because each group during training contains at least one positive and one negative sample, the analysis considers the gap in both positive and negative samples. In addition, the ratio of the negative-to-positive gap is examined. The logit gap and ratio are defined as follows.

\begin{equation} \label{eq:gap}
    g_i = \frac{1}{T_i}
\sum_{t=1}^{T_i}
\left(
\max_{v \in \mathcal{V}} z_{t,v}
-
z_{t,y_t}
\right)
\end{equation}
 and 
\begin{equation} \label{eq:ratio}
    \mathrm{Gap Ratio}
=
\frac{\bar{g}_{\mathrm{neg}}}
{\bar{g}_{\mathrm{pos}}+\epsilon} 
\end{equation}

where $T_i$ denotes the number of code tokens in sample $i$, $\mathcal{V}$ is the vocabulary, $z_{t,v}$ is the logit assigned to token $v$ at position $t$, and $y_t$ is the corresponding token from the offline dataset. $\bar{g}_{\mathrm{pos}}$ and $\bar{g}_{\mathrm{neg}}$ denote the average sequence-level logit gap over positive (reward $=1$) and negative (reward $<0$) samples, respectively. A small constant $\epsilon = 10^{-8}$ is added for numerical stability.

Figure \ref{fig:logit_gap_qwen} presents the plots of positive gap, negative gap, and gap ratio for Qwen 7B (see Appendix \ref{sec:logit_gap} for other models). The initial gap is close to zero, supporting the hypothesis outlined in Section \ref{hypothesis}. The pretrained large language model (LLM) generally assigns the highest probability to tokens from the dataset. As training progresses, the logit gap increases.

Because the model is trained to favor positive samples and disfavor negative samples, an increase in the negative gap is expected. However, the positive gap also increases. Alongside improved performance, this increase in the positive gap indicates that the model not only learns to generate correct code, but also develops a preference for certain positive samples over others. This behavior resembles that observed in online RL training.

The gap ratio demonstrates that although both positive and negative gaps increase, the negative gap typically grows more rapidly. When both gaps increase proportionally, as observed with a low gap ratio (e.g., learning rate = $5\times10^{-5}$), performance does not improve. In contrast, a high gap ratio, characterized by a substantial negative gap, results in performance collapse. This outcome may occur because the model stops learning from negative samples. If the negative gap is high early in training (learning rate = $2\times10^{-5}$), the performance of the model remains poor. In contrast, if the negative gap starts low and then increases (learning rate = $1\times10^{-5}$), the model shows rapid improvement before eventual degradation.

Analysis of logit gaps and ratios in the Qwen 7B model indicates that an increase in the negative gap is a primary factor that contributes to performance collapse. Incorporating new samples from the dataset or sampling additional negative examples from the model after each epoch can help prevent collapse and support continued model improvement. The introduction of new samples is briefly examined in Appendix \ref{resampling}, where an improvement in model performance is observed. Developing a comprehensive resampling strategy remains an area for future research.

\section{Related Work}
Post-training with RL is commonly conducted using online sampling of the model under training. However, because transformers are computationally inefficient, an efficient approximation is used for sampling \cite{yao2025offpolicy}, which renders training off-policy in most practical settings. Training typically uses GRPO \cite{grpo} with verifiable rewards \cite{rlvr}.
\citet{mugrpo, asyncrl, offpolicyrl, coderl} investigated the application of GRPO in off-policy online settings, where data are sampled from the model asynchronously with respect to the training loop. 

In contrast, this work trains the model without any sampling. 
Furthermore, previous works have studied off-policy RL for Maths, while this study examines offline RL in the context of code generation, where achieving a correct final answer is insufficient, and the model must also acquire syntactic understanding. 

\citet{ILQL} formulated offline RL as a partially observable Markov decision process and employed Q-learning for training. ILQL suffers from instability during inference when extracting policy from value heads. However, the reasons for instability are not explored. \citet{offline} demonstrated that offline RL can serve as a viable alternative to online RL. However, their work did not include a comprehensive analysis of model families, hyperparameters, or causes of failure in offline settings. Unlike other studies, this work is the first to analyze the performance and stability of policy gradient algorithms directly for offline RL without sampling the model being trained. 

Additional related works are discussed in Appendix \ref{more related work}.

\section{Conclusion}

This work investigates complete offline post-training of coding LLMs using samples and rewards derived from an existing dataset, rather than those generated by an LLM. The results demonstrate that offline RL can substantially enhance the performance of the base LLM after only a few hours of training on a single GPU. These improvements are consistent across different families and sizes of models. Nevertheless, the overall training dynamics remains unstable. The analysis identifies logit variance and the logit gap between tokens favored by the model and those present in the offline dataset as primary sources of instability. Our aim with this work is to take a step towards investigating the feasibility of offline RL and to identify its failure modes. 
Implementing methods to control these sources of failure, as well as incorporating samples generated by the model during training, can further improve model performance.

\section*{Limitations}
We analyze offline post-training for LLMs for code generation. We show that offline RL can be unstable. While our analysis points towards different mechanisms to control the instability, we do not verify all the benefits of these mechanisms and leave it for future work. Since our aim in the work was to establish the feasibility of offline RL and not to have a new state-of-the-art code model, we have only performed an evaluation on two benchmarks. The evaluation is sufficient to show that offline RL can improve the model. We have only trained the model with code from a single language (Python). Thus, our work does not establish whether offline RL works on generating code for multiple programming languages.
Our analysis reveals that although the causes of instability in both online and offline RL are different, the impact (collapse) remains the same \cite{offpolicyrl}. By adapting methods used to stabilize online training to the offline RL setup can lead to stable offline RL training. We leave such exploration for future work. 

\section*{Ethical Considerations}
In this work we have studied Offline RL for LLM as an efficient alternative to Online RL. Our objective is to reduce the computational and energy resources required to align LLMs to human preferences (functional correctness of code in this work). However, it is possible to use the same methodology to train LLMs to generate harmful contents using very few computational resources. We acknowledge this risk associated with the development of efficient RL training.


\bibliography{custom}

\begin{thebibliography}{32}
\providecommand{\natexlab}[1]{#1}

\bibitem[{Ahmadian et~al.(2024)Ahmadian, Cremer, Gallé, Fadaee, Kreutzer, Pietquin, Üstün, and Hooker}]{rloo}
Arash Ahmadian, Chris Cremer, Matthias Gallé, Marzieh Fadaee, Julia Kreutzer, Olivier Pietquin, Ahmet Üstün, and Sara Hooker. 2024.
\newblock \href {https://arxiv.org/abs/2402.14740} {Back to basics: Revisiting reinforce style optimization for learning from human feedback in llms}.
\newblock \emph{Preprint}, arXiv:2402.14740.

\bibitem[{Anand et~al.(2026)Anand, mingze wu, Verma, and Mezini}]{offline}
Abhinav Anand, mingze wu, Shweta Verma, and Mira Mezini. 2026.
\newblock \href {https://openreview.net/forum?id=HyxnU8GJmR} {Efficient post-training of {LLM}s for code generation with offline reinforcement learning}.
\newblock In \emph{Decision-Making from Offline Datasets to Online Adaptation: Black-Box Optimization to Reinforcement Learning}.

\bibitem[{Andrew and Gao(2007)}]{andrew2007scalable}
Galen Andrew and Jianfeng Gao. 2007.
\newblock Scalable training of {L1}-regularized log-linear models.
\newblock In \emph{Proceedings of the 24th International Conference on Machine Learning}, pages 33--40.

\bibitem[{Austin et~al.(2021)Austin, Odena, Nye, Bosma, Michalewski, Dohan, Jiang, Cai, Terry, Le, and Sutton}]{mbpp}
Jacob Austin, Augustus Odena, Maxwell Nye, Maarten Bosma, Henryk Michalewski, David Dohan, Ellen Jiang, Carrie Cai, Michael Terry, Quoc Le, and Charles Sutton. 2021.
\newblock \href {https://arxiv.org/abs/2108.07732} {Program synthesis with large language models}.
\newblock \emph{Preprint}, arXiv:2108.07732.

\bibitem[{Fu et~al.(2026)Fu, Gao, Shen, Zhu, Mei, He, Xu, Wei, Mei, Wang, Yang, Yuan, and Wu}]{asyncrl}
Wei Fu, Jiaxuan Gao, Xujie Shen, Chen Zhu, Zhiyu Mei, Chuyi He, Shusheng Xu, Guo Wei, Jun Mei, Jiashu Wang, Tongkai Yang, Binhang Yuan, and Yi~Wu. 2026.
\newblock \href {https://arxiv.org/abs/2505.24298} {Areal: A large-scale asynchronous reinforcement learning system for language reasoning}.
\newblock \emph{Preprint}, arXiv:2505.24298.

\bibitem[{Guo et~al.(2025)Guo, Yang, Zhang, Song, Wang, Zhu, Xu, Zhang, Ma, Bi, Zhang, Yu, Wu, Wu, Gou, Shao, Li, Gao, Liu, Xue, Wang, Wu, Feng, Lu, Zhao, Deng, Ruan, Dai, Chen, Ji, Li, Lin, Dai, Luo, Hao, Chen, Li, Zhang, Xu, Ding, Gao, Qu, Li, Guo, Li, Chen, Yuan, Tu, Qiu, Li, Cai, Ni, Liang, Chen, Dong, Hu, You, Gao, Guan, Huang, Yu, Wang, Zhang, Zhao, Wang, Zhang, Xu, Xia, Zhang, Zhang, Tang, Zhou, Li, Wang, Li, Tian, Huang, Zhang, Wang, Chen, Du, Ge, Zhang, Pan, Wang, Chen, Jin, Chen, Lu, Zhou, Chen, Ye, Wang, Yu, Zhou, Pan, Li, Zhou, Wu, Yun, Pei, Sun, Wang, Zeng, Liu, Liang, Gao, Yu, Zhang, Xiao, An, Liu, Wang, Chen, Nie, Cheng, Liu, Xie, Liu, Yang, Li, Su, Lin, Li, Jin, Shen, Chen, Sun, Wang, Song, Zhou, Wang, Shan, Li, Wang, Wei, Zhang, Xu, Li, Zhao, Sun, Wang, Yu, Zhang, Shi, Xiong, He, Piao, Wang, Tan, Ma, Liu, Guo, Ou, Wang, Gong, Zou, He, Xiong, Luo, You, Liu, Zhou, Zhu, Huang, Li, Zheng, Zhu, Ma, Tang, Zha, Yan, Ren, Ren, Sha, Fu, Xu, Xie, Zhang, Hao, Ma, Yan, Wu, Gu, Zhu, Liu, Li, Xie, Song,
  Pan, Huang, Xu, Zhang, and Zhang}]{deepseekr1}
Daya Guo, Dejian Yang, Haowei Zhang, Junxiao Song, Peiyi Wang, Qihao Zhu, Runxin Xu, Ruoyu Zhang, Shirong Ma, Xiao Bi, Xiaokang Zhang, Xingkai Yu, Yu~Wu, Z.~F. Wu, Zhibin Gou, Zhihong Shao, Zhuoshu Li, Ziyi Gao, Aixin Liu, and 175 others. 2025.
\newblock \href {https://doi.org/10.1038/s41586-025-09422-z} {Deepseek-r1 incentivizes reasoning in llms through reinforcement learning}.
\newblock \emph{Nature}, 645(8081):633–638.

\bibitem[{Guo et~al.(2024)Guo, Zhu, Yang, Xie, Dong, Zhang, Chen, Bi, Wu, Li, Luo, Xiong, and Liang}]{deepseekcoder}
Daya Guo, Qihao Zhu, Dejian Yang, Zhenda Xie, Kai Dong, Wentao Zhang, Guanting Chen, Xiao Bi, Y.~Wu, Y.~K. Li, Fuli Luo, Yingfei Xiong, and Wenfeng Liang. 2024.
\newblock \href {https://arxiv.org/abs/2401.14196} {Deepseek-coder: When the large language model meets programming -- the rise of code intelligence}.
\newblock \emph{Preprint}, arXiv:2401.14196.

\bibitem[{Han et~al.(2025)Han, You, Wang, Luo, Yang, Shi, Chen, Zhang, Lan, Deng, Ji, Liu, Huang, Zhang, Pan, Wang, Huang, Li, and Wu}]{asyncflow}
Zhenyu Han, Ansheng You, Haibo Wang, Kui Luo, Guang Yang, Wenqi Shi, Menglong Chen, Sicheng Zhang, Zeshun Lan, Chunshi Deng, Huazhong Ji, Wenjie Liu, Yu~Huang, Yixiang Zhang, Chenyi Pan, Jing Wang, Xin Huang, Chunsheng Li, and Jianping Wu. 2025.
\newblock \href {https://arxiv.org/abs/2507.01663} {Asyncflow: An asynchronous streaming rl framework for efficient llm post-training}.
\newblock \emph{Preprint}, arXiv:2507.01663.

\bibitem[{Hendrycks et~al.(2021)Hendrycks, Basart, Kadavath, Mazeika, Arora, Guo, Burns, Puranik, He, Song, and Steinhardt}]{apps}
Dan Hendrycks, Steven Basart, Saurav Kadavath, Mantas Mazeika, Akul Arora, Ethan Guo, Collin Burns, Samir Puranik, Horace He, Dawn Song, and Jacob Steinhardt. 2021.
\newblock \href {https://arxiv.org/abs/2105.09938} {Measuring coding challenge competence with apps}.
\newblock \emph{Preprint}, arXiv:2105.09938.

\bibitem[{Hui et~al.(2024)Hui, Yang, Cui, Yang, Liu, Zhang, Liu, Zhang, Yu, Lu, Dang, Fan, Zhang, Yang, Men, Huang, Zheng, Miao, Quan, Feng, Ren, Ren, Zhou, and Lin}]{qwencoder}
Binyuan Hui, Jian Yang, Zeyu Cui, Jiaxi Yang, Dayiheng Liu, Lei Zhang, Tianyu Liu, Jiajun Zhang, Bowen Yu, Keming Lu, Kai Dang, Yang Fan, Yichang Zhang, An~Yang, Rui Men, Fei Huang, Bo~Zheng, Yibo Miao, Shanghaoran Quan, and 5 others. 2024.
\newblock \href {https://arxiv.org/abs/2409.12186} {Qwen2.5-coder technical report}.
\newblock \emph{Preprint}, arXiv:2409.12186.

\bibitem[{Jiang et~al.(2026)Jiang, Wang, Shen, Kim, and Kim}]{cllm_survey}
Juyong Jiang, Fan Wang, Jiasi Shen, Sungju Kim, and Sunghun Kim. 2026.
\newblock \href {https://doi.org/10.1145/3747588} {A survey on large language models for code generation}.
\newblock \emph{ACM Transactions on Software Engineering and Methodology}, 35(2):1–72.

\bibitem[{Kocetkov et~al.(2022)Kocetkov, Li, Allal, Li, Mou, Ferrandis, Jernite, Mitchell, Hughes, Wolf, Bahdanau, von Werra, and de~Vries}]{stack}
Denis Kocetkov, Raymond Li, Loubna~Ben Allal, Jia Li, Chenghao Mou, Carlos~Muñoz Ferrandis, Yacine Jernite, Margaret Mitchell, Sean Hughes, Thomas Wolf, Dzmitry Bahdanau, Leandro von Werra, and Harm de~Vries. 2022.
\newblock \href {https://arxiv.org/abs/2211.15533} {The stack: 3 tb of permissively licensed source code}.
\newblock \emph{Preprint}, arXiv:2211.15533.

\bibitem[{Kool et~al.(2019)Kool, van Hoof, and Welling}]{rloobuy}
Wouter Kool, Herke van Hoof, and Max Welling. 2019.
\newblock \href {https://openreview.net/forum?id=r1lgTGL5DE} {Buy 4 {REINFORCE} samples, get a baseline for free!}

\bibitem[{Kwon et~al.(2023)Kwon, Li, Zhuang, Sheng, Zheng, Yu, Gonzalez, Zhang, and Stoica}]{vllm}
Woosuk Kwon, Zhuohan Li, Siyuan Zhuang, Ying Sheng, Lianmin Zheng, Cody~Hao Yu, Joseph~E. Gonzalez, Hao Zhang, and Ion Stoica. 2023.
\newblock Efficient memory management for large language model serving with pagedattention.
\newblock In \emph{Proceedings of the ACM SIGOPS 29th Symposium on Operating Systems Principles}.

\bibitem[{Lambert et~al.(2025)Lambert, Morrison, Pyatkin, Huang, Ivison, Brahman, Miranda, Liu, Dziri, Lyu, Gu, Malik, Graf, Hwang, Yang, Bras, Tafjord, Wilhelm, Soldaini, Smith, Wang, Dasigi, and Hajishirzi}]{rlvr}
Nathan Lambert, Jacob Morrison, Valentina Pyatkin, Shengyi Huang, Hamish Ivison, Faeze Brahman, Lester James~V. Miranda, Alisa Liu, Nouha Dziri, Shane Lyu, Yuling Gu, Saumya Malik, Victoria Graf, Jena~D. Hwang, Jiangjiang Yang, Ronan~Le Bras, Oyvind Tafjord, Chris Wilhelm, Luca Soldaini, and 4 others. 2025.
\newblock \href {https://arxiv.org/abs/2411.15124} {Tulu 3: Pushing frontiers in open language model post-training}.
\newblock \emph{Preprint}, arXiv:2411.15124.

\bibitem[{Le et~al.(2022)Le, Wang, Gotmare, Savarese, and Hoi}]{coderl}
Hung Le, Yue Wang, Akhilesh~Deepak Gotmare, Silvio Savarese, and Steven C.~H. Hoi. 2022.
\newblock \href {https://arxiv.org/abs/2207.01780} {Coderl: Mastering code generation through pretrained models and deep reinforcement learning}.
\newblock \emph{Preprint}, arXiv:2207.01780.

\bibitem[{Levine et~al.(2020)Levine, Kumar, Tucker, and Fu}]{offlinetut}
Sergey Levine, Aviral Kumar, George Tucker, and Justin Fu. 2020.
\newblock \href {https://arxiv.org/abs/2005.01643} {Offline reinforcement learning: Tutorial, review, and perspectives on open problems}.
\newblock \emph{Preprint}, arXiv:2005.01643.

\bibitem[{Lozhkov et~al.(2024)Lozhkov, Li, Allal, Cassano, Lamy-Poirier, Tazi, Tang, Pykhtar, Liu, Wei, Liu, Tian, Kocetkov, Zucker, Belkada, Wang, Liu, Abulkhanov, Paul, Li, Li, Risdal, Li, Zhu, Zhuo, Zheltonozhskii, Dade, Yu, Krauß, Jain, Su, He, Dey, Abati, Chai, Muennighoff, Tang, Oblokulov, Akiki, Marone, Mou, Mishra, Gu, Hui, Dao, Zebaze, Dehaene, Patry, Xu, McAuley, Hu, Scholak, Paquet, Robinson, Anderson, Chapados, Patwary, Tajbakhsh, Jernite, Ferrandis, Zhang, Hughes, Wolf, Guha, von Werra, and de~Vries}]{stackv2}
Anton Lozhkov, Raymond Li, Loubna~Ben Allal, Federico Cassano, Joel Lamy-Poirier, Nouamane Tazi, Ao~Tang, Dmytro Pykhtar, Jiawei Liu, Yuxiang Wei, Tianyang Liu, Max Tian, Denis Kocetkov, Arthur Zucker, Younes Belkada, Zijian Wang, Qian Liu, Dmitry Abulkhanov, Indraneil Paul, and 47 others. 2024.
\newblock \href {https://arxiv.org/abs/2402.19173} {Starcoder 2 and the stack v2: The next generation}.
\newblock \emph{Preprint}, arXiv:2402.19173.

\bibitem[{Noukhovitch et~al.(2025)Noukhovitch, Huang, Xhonneux, Hosseini, Agarwal, and Courville}]{asyncrlhf}
Michael Noukhovitch, Shengyi Huang, Sophie Xhonneux, Arian Hosseini, Rishabh Agarwal, and Aaron Courville. 2025.
\newblock \href {https://arxiv.org/abs/2410.18252} {Asynchronous rlhf: Faster and more efficient off-policy rl for language models}.
\newblock \emph{Preprint}, arXiv:2410.18252.

\bibitem[{Puri et~al.(2021)Puri, Kung, Janssen, Zhang, Domeniconi, Zolotov, Dolby, Chen, Choudhury, Decker, Thost, Buratti, Pujar, Ramji, Finkler, Malaika, and Reiss}]{codenet}
Ruchir Puri, David~S. Kung, Geert Janssen, Wei Zhang, Giacomo Domeniconi, Vladimir Zolotov, Julian Dolby, Jie Chen, Mihir Choudhury, Lindsey Decker, Veronika Thost, Luca Buratti, Saurabh Pujar, Shyam Ramji, Ulrich Finkler, Susan Malaika, and Frederick Reiss. 2021.
\newblock \href {https://arxiv.org/abs/2105.12655} {Codenet: A large-scale ai for code dataset for learning a diversity of coding tasks}.
\newblock \emph{Preprint}, arXiv:2105.12655.

\bibitem[{Rozière et~al.(2024)Rozière, Gehring, Gloeckle, Sootla, Gat, Tan, Adi, Liu, Sauvestre, Remez, Rapin, Kozhevnikov, Evtimov, Bitton, Bhatt, Ferrer, Grattafiori, Xiong, Défossez, Copet, Azhar, Touvron, Martin, Usunier, Scialom, and Synnaeve}]{codellama}
Baptiste Rozière, Jonas Gehring, Fabian Gloeckle, Sten Sootla, Itai Gat, Xiaoqing~Ellen Tan, Yossi Adi, Jingyu Liu, Romain Sauvestre, Tal Remez, Jérémy Rapin, Artyom Kozhevnikov, Ivan Evtimov, Joanna Bitton, Manish Bhatt, Cristian~Canton Ferrer, Aaron Grattafiori, Wenhan Xiong, Alexandre Défossez, and 7 others. 2024.
\newblock \href {https://arxiv.org/abs/2308.12950} {Code llama: Open foundation models for code}.
\newblock \emph{Preprint}, arXiv:2308.12950.

\bibitem[{Schulman et~al.(2017)Schulman, Wolski, Dhariwal, Radford, and Klimov}]{ppo}
John Schulman, Filip Wolski, Prafulla Dhariwal, Alec Radford, and Oleg Klimov. 2017.
\newblock \href {https://arxiv.org/abs/1707.06347} {Proximal policy optimization algorithms}.
\newblock \emph{Preprint}, arXiv:1707.06347.

\bibitem[{Shao et~al.(2024)Shao, Wang, Zhu, Xu, Song, Bi, Zhang, Zhang, Li, Wu, and Guo}]{grpo}
Zhihong Shao, Peiyi Wang, Qihao Zhu, Runxin Xu, Junxiao Song, Xiao Bi, Haowei Zhang, Mingchuan Zhang, Y.~K. Li, Y.~Wu, and Daya Guo. 2024.
\newblock \href {https://arxiv.org/abs/2402.03300} {Deepseekmath: Pushing the limits of mathematical reasoning in open language models}.
\newblock \emph{Preprint}, arXiv:2402.03300.

\bibitem[{Sheng et~al.(2025)Sheng, Zhang, Ye, Wu, Zhang, Zhang, Peng, Lin, and Wu}]{verl}
Guangming Sheng, Chi Zhang, Zilingfeng Ye, Xibin Wu, Wang Zhang, Ru~Zhang, Yanghua Peng, Haibin Lin, and Chuan Wu. 2025.
\newblock \href {https://doi.org/10.1145/3689031.3696075} {Hybridflow: A flexible and efficient rlhf framework}.
\newblock In \emph{Proceedings of the Twentieth European Conference on Computer Systems}, EuroSys ’25, page 1279–1297. ACM.

\bibitem[{Shojaee et~al.(2023)Shojaee, Jain, Tipirneni, and Reddy}]{ppocoder}
Parshin Shojaee, Aneesh Jain, Sindhu Tipirneni, and Chandan~K. Reddy. 2023.
\newblock \href {https://arxiv.org/abs/2301.13816} {Execution-based code generation using deep reinforcement learning}.
\newblock \emph{Preprint}, arXiv:2301.13816.

\bibitem[{Snell et~al.(2023)Snell, Kostrikov, Su, Yang, and Levine}]{ILQL}
Charlie Snell, Ilya Kostrikov, Yi~Su, Mengjiao Yang, and Sergey Levine. 2023.
\newblock \href {https://arxiv.org/abs/2206.11871} {Offline rl for natural language generation with implicit language q learning}.
\newblock \emph{Preprint}, arXiv:2206.11871.

\bibitem[{Tian et~al.(2026)Tian, Xie, and Wei}]{mugrpo}
Minghao Tian, Yunfei Xie, and Chen Wei. 2026.
\newblock \href {https://arxiv.org/abs/2605.17570} {How off-policy can grpo be? mu-grpo for efficient llm reinforcement learning}.
\newblock \emph{Preprint}, arXiv:2605.17570.

\bibitem[{Vaswani et~al.(2023)Vaswani, Shazeer, Parmar, Uszkoreit, Jones, Gomez, Kaiser, and Polosukhin}]{attention}
Ashish Vaswani, Noam Shazeer, Niki Parmar, Jakob Uszkoreit, Llion Jones, Aidan~N. Gomez, Lukasz Kaiser, and Illia Polosukhin. 2023.
\newblock \href {https://arxiv.org/abs/1706.03762} {Attention is all you need}.
\newblock \emph{Preprint}, arXiv:1706.03762.

\bibitem[{Williams(1992)}]{reinforce}
Ronald~J. Williams. 1992.
\newblock \href {https://doi.org/10.1007/BF00992696} {Simple statistical gradient-following algorithms for connectionist reinforcement learning}.
\newblock \emph{Mach. Learn.}, 8(3–4):229–256.

\bibitem[{Yao et~al.(2025)Yao, Liu, Zhang, Dong, Shang, and Gao}]{yao2025offpolicy}
Feng Yao, Liyuan Liu, Dinghuai Zhang, Chengyu Dong, Jingbo Shang, and Jianfeng Gao. 2025.
\newblock \href {https://fengyao.notion.site/off-policy-rl} {Your efficient rl framework secretly brings you off-policy rl training}.

\bibitem[{Zheng et~al.(2026)Zheng, Zhao, and Chen}]{offpolicyrl}
Haizhong Zheng, Jiawei Zhao, and Beidi Chen. 2026.
\newblock \href {https://openreview.net/forum?id=IIgl5MWelz} {Prosperity before collapse: How far can off-policy {RL} reach with stale data on {LLM}s?}
\newblock In \emph{The Fourteenth International Conference on Learning Representations}.

\bibitem[{Zhong et~al.(2025)Zhong, Zhang, Song, Hu, Jin, Wu, Chen, Chen, Zhou, Wan, Zhou, Jiang, Zhu, and Jiang}]{streamrl}
Yinmin Zhong, Zili Zhang, Xiaoniu Song, Hanpeng Hu, Chao Jin, Bingyang Wu, Nuo Chen, Yukun Chen, Yu~Zhou, Changyi Wan, Hongyu Zhou, Yimin Jiang, Yibo Zhu, and Daxin Jiang. 2025.
\newblock \href {https://arxiv.org/abs/2504.15930} {Streamrl: Scalable, heterogeneous, and elastic rl for llms with disaggregated stream generation}.
\newblock \emph{Preprint}, arXiv:2504.15930.

\end{thebibliography}

\appendix
\section{More Related Work} \label{more related work}
LLM post-training is usually performed using Online RL algorithms such as PPO \cite{ppo}, GRPO \cite{grpo} or RLOO \cite{rloo}. Among these, GRPO is the most commonly used algorithm as it provides the best balance between performance and resource requirement.  The training is typically performed using RL optimization libraries such as verl \cite{verl}. To ensure high throughput during inference, libraries such as SGLang or vLLM \cite{vllm} are employed. However, the use of these libraries for inference instead of using the full model being trained makes the post-training off-policy \cite{yao2025offpolicy}.

Due to the practical off-policy nature of post-training, many works have studied how off-policy post-training can be \cite{offpolicyrl, mugrpo, andrew2007scalable}. These works propose different methodology for efficient asynchronous sampling \cite{asyncflow, streamrl, asyncrlhf}. In contrast, we perform a complete offline post-training where no sampling and verification takes place. Instead, both code samples and their execution status comes from pre-existing dataset.

Similar to our work, off-policy RL also leads to model collapse due to staleness \cite{offpolicyrl}. Different strategies have been proposed to stabilize off-policy RL \cite{offpolicyrl, mugrpo}. These strategies are complementary to our work and can also help in stabilizing  offline RL post-training.

\section{RL Algorithm}\label{RLOO}
RLOO is a policy optimization algorithm that reduces the variance of policy-gradient updates without the need for a separate value function. Rather than estimating a baseline value, RLOO calculates the advantage of each response relative to the rewards obtained by other responses generated for the same input. This baseline improves the stability of the training without requiring an additional value network.

Given a group of $K$ responses with rewards $\{r_1,r_2,\ldots,r_K\}$, the leave-one-out advantage for the $i$-th response is computed as follows:

\begin{equation}
A_i = r_i - \frac{1}{K-1}\sum_{j\neq i} r_j,
\end{equation}

where $r_i$ denotes the reward assigned to the $i$-th response. Responses with rewards exceeding the group average receive positive advantages, whereas those with lower rewards receive negative advantages. These computed advantages determine the direction and magnitude of the policy update, promoting higher probabilities for better-performing responses. Compared with methods that utilize a learned value function, RLOO offers a simpler optimization process and reduced memory requirements, making it particularly suitable for LLMs.

\section{Performance on APPS}
\begin{figure*}
    
    \begin{subfigure}[b]{0.99\textwidth}
     
     \includegraphics[width=\linewidth]{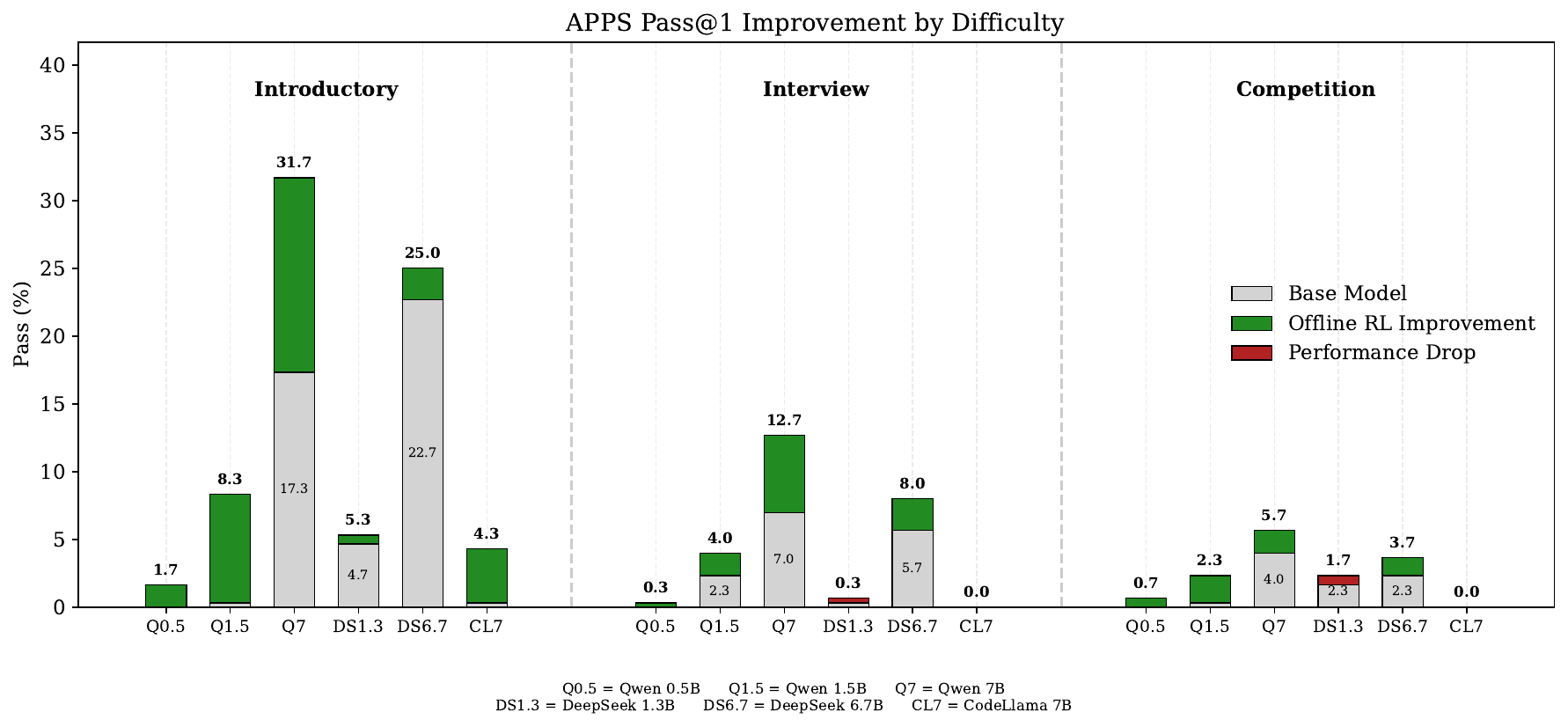}
    \end{subfigure}

    \begin{subfigure}[b]{0.99\textwidth}
      
     \includegraphics[width=\linewidth]{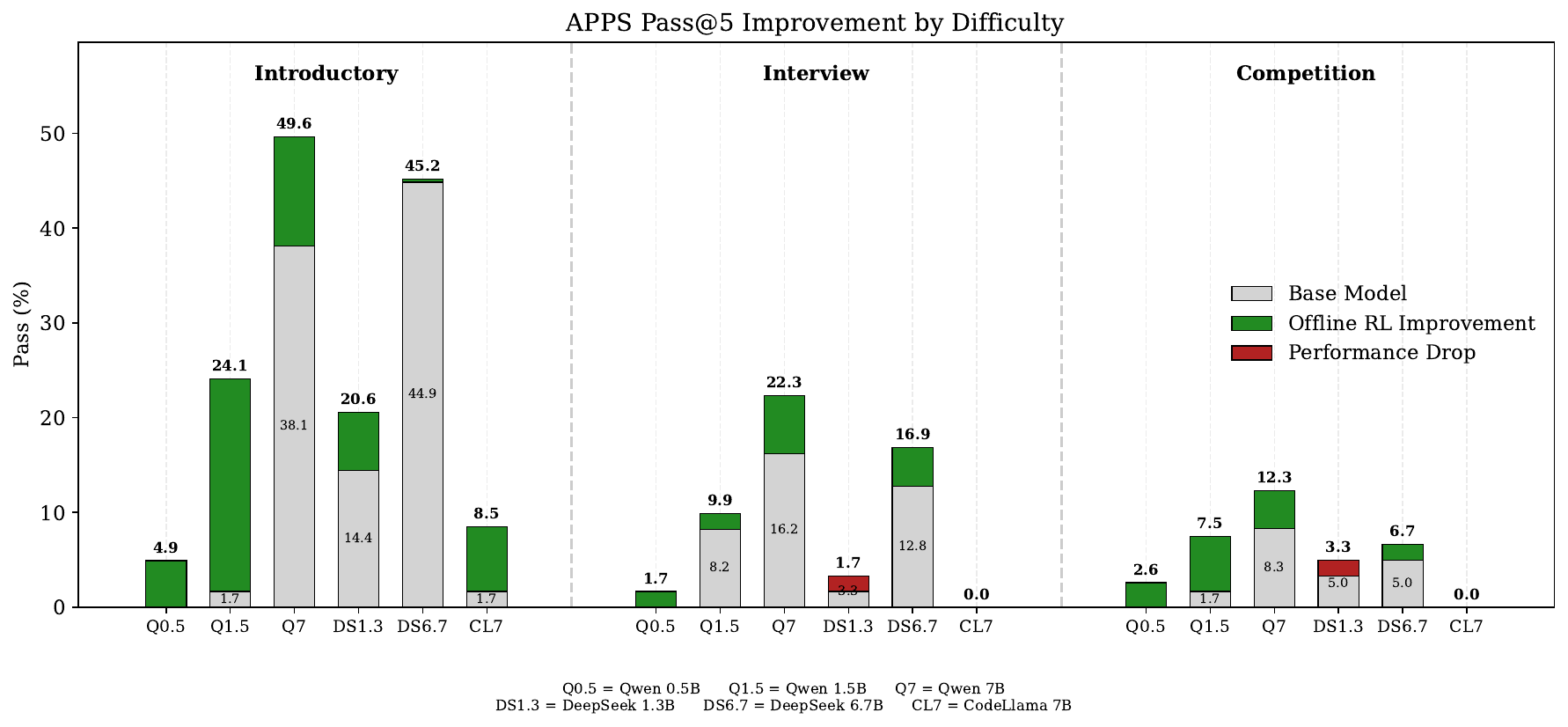} 
    \end{subfigure}

    \begin{subfigure}[b]{0.99\textwidth}
      
     \includegraphics[width=\linewidth]{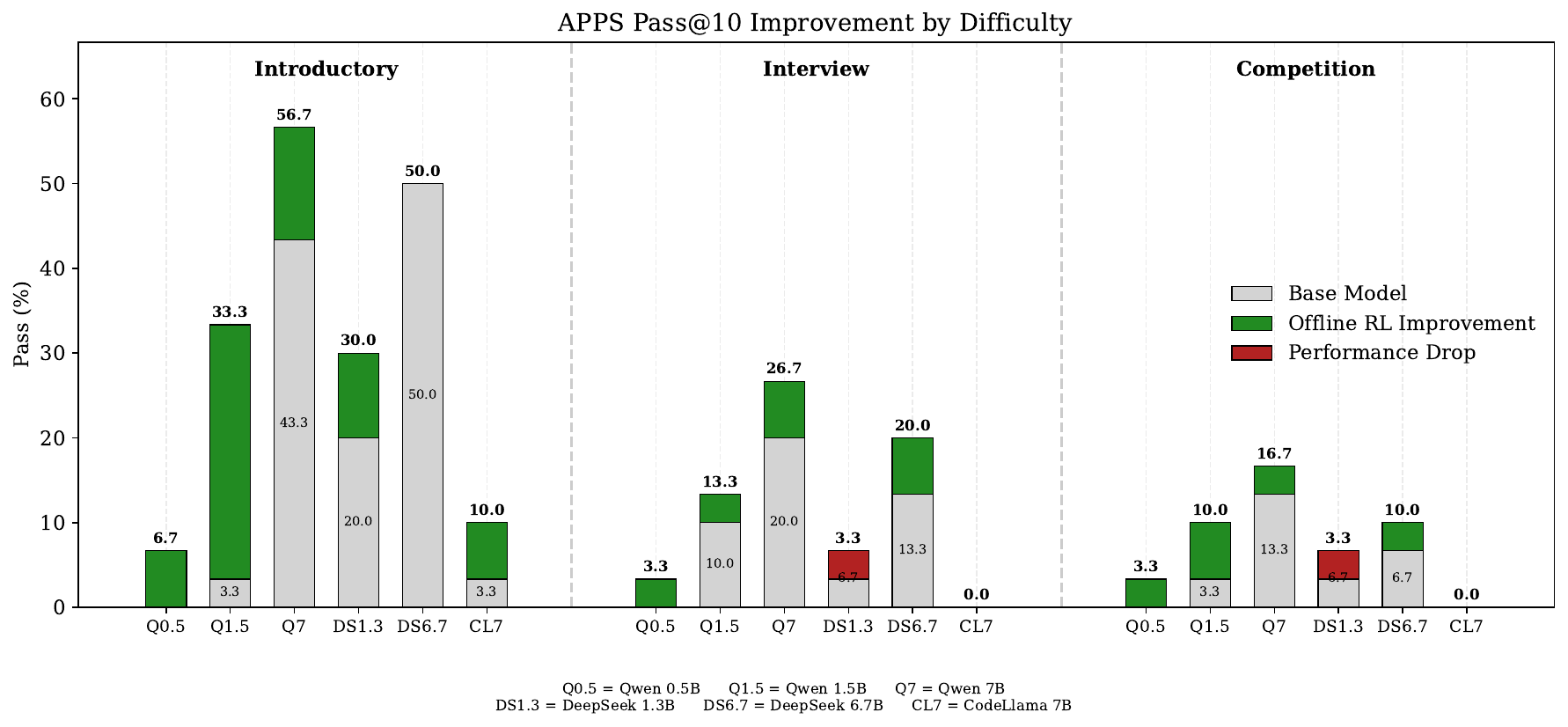} 
    \end{subfigure}
    
\caption{Performance in terms of pass@1, pass@5 and pass@10 on APPS dataset across different difficulty levels.}
\label{fig:apps_difficulty}
\end{figure*}

Figure \ref{fig:apps_difficulty} shows the performance of all models on the different difficulty level of APPS dataset. All models, except DeepSeek-1.3B improves performance across all difficulty levels. Qwen models achieve the best improvement over the base model.

\begin{table*}[t]
\centering
\small
\begin{tabular}{lccccc}
\hline
\textbf{Model} & \textbf{Best LR} & \textbf{Base Model Pass@1 (\%)}& \textbf{Peak Pass@1 (\%)} & \textbf{Epoch} & \textbf{Final Pass@1 (\%)} \\
\hline
Qwen 0.5B      & 3e-5 & 0 &48 & 4 & 2   \\
Qwen 1.5B      & 3e-5 & 13 &54 & 2 & 0   \\
DeepSeek 1.3B  & 2e-6 & 56 &57 & 5 & 57  \\
Qwen 7B        & 1e-5 & 64 &79 & 1 & 0    \\
DeepSeek 6.7B  & 2e-5 & 3  &34 & 4 & 11  \\
CodeLlama 7B     & 3e-5 & 0 &9  & 1 & 0   \\
\hline
\end{tabular}
\caption{Summary of MBPP Pass@1 results for the configurations achieving the highest sustained performance within each model family.}
\label{tab:performance_summary}
\end{table*}

\setlength{\tabcolsep}{4pt}

\begin{figure*}[h]
\centering
    \begin{subfigure}[b]{0.99\textwidth}
     \includegraphics[width=0.33\linewidth]{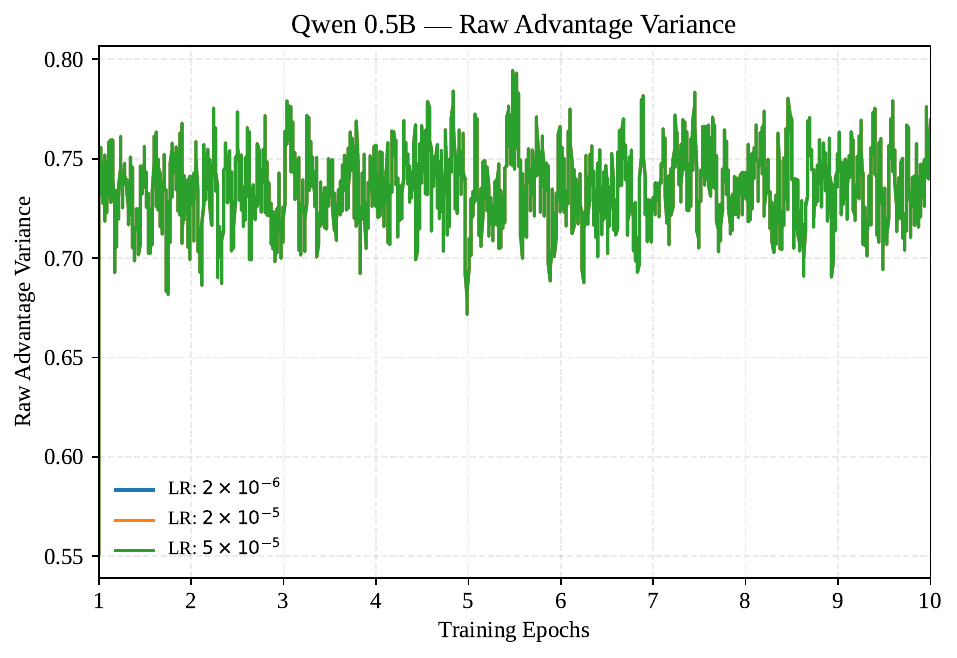} 
     \includegraphics[width=0.33\linewidth]{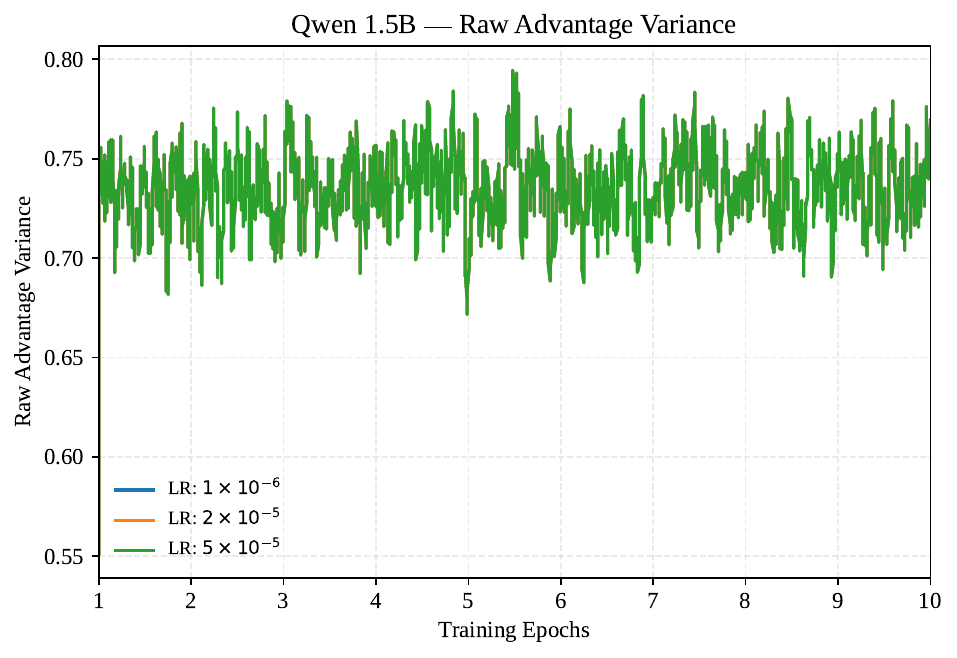}
     \includegraphics[width=0.33\linewidth]{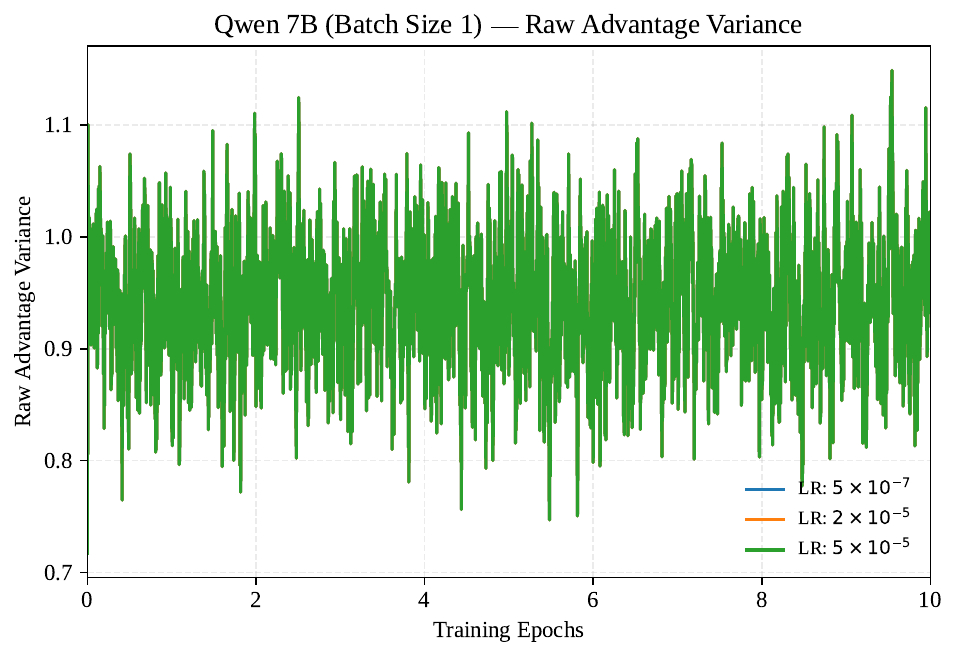}
    \end{subfigure}

    \begin{subfigure}[b]{0.99\textwidth}
     \includegraphics[width=0.33\linewidth]{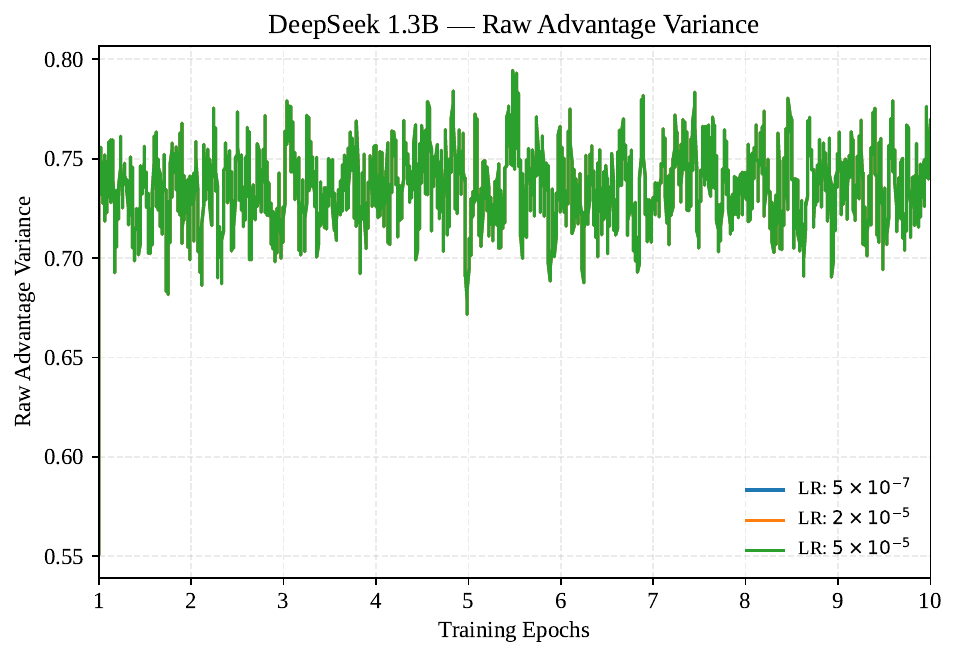} 
     \includegraphics[width=0.33\linewidth]{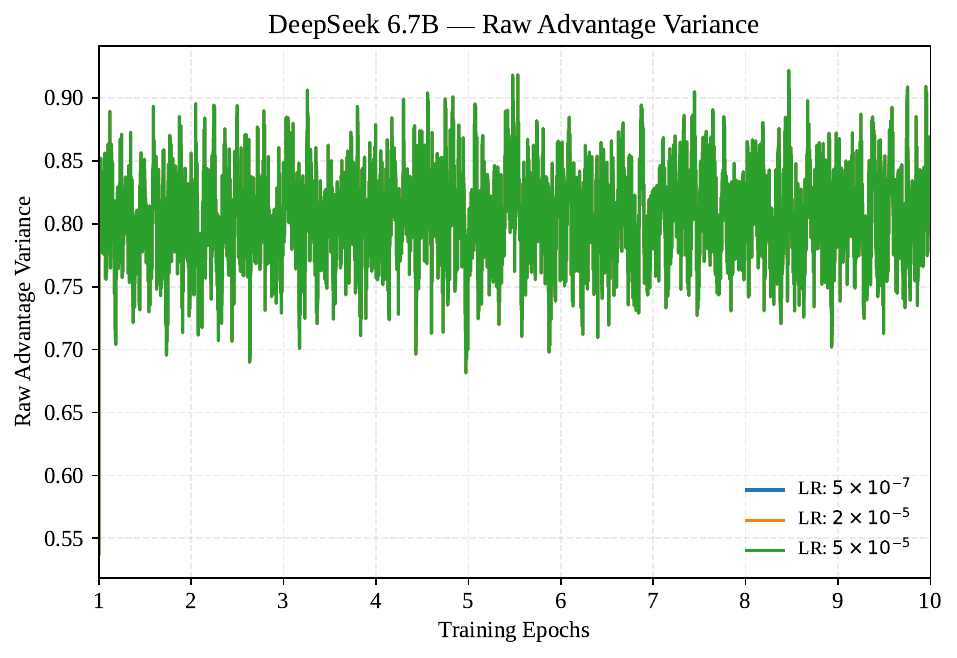}
     \includegraphics[width=0.33\linewidth]{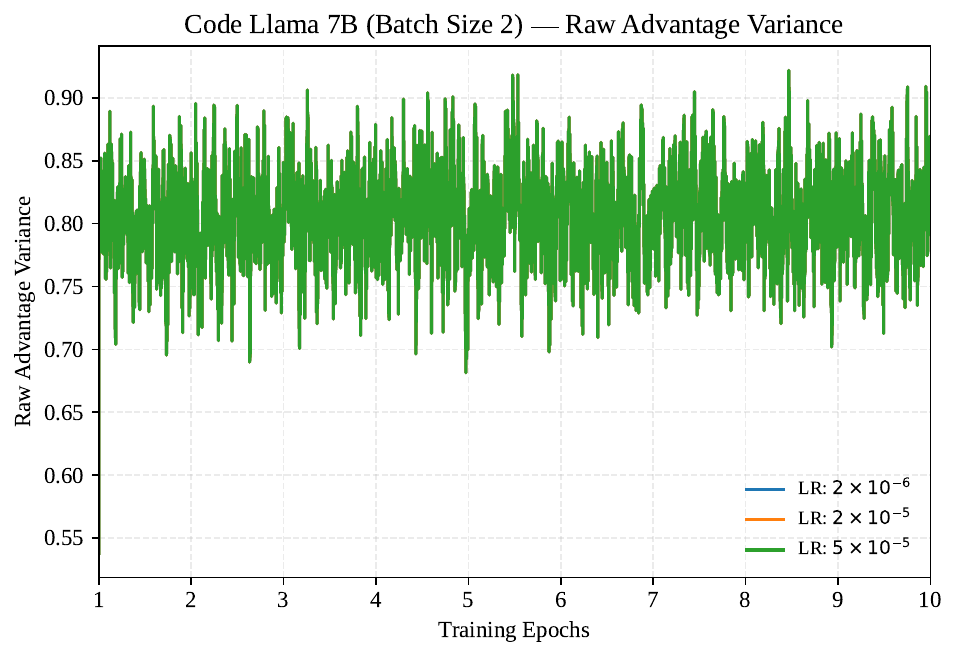} 
    \end{subfigure}
    
\caption{Advantage variance across training epochs for all models. Within each batch size group, the curves across all evaluated learning rates overlap almost perfectly demonstrating that advantage variance remains completely invariant to optimization settings.}
\label{fig:raw_adv_variance}
\end{figure*}

\section{Performance on MBPP}
Table \ref{tab:performance_summary} shows the pass@1 on the MBPP dataset for the base model and with offline RL training. The table shows that different models achieve peak performance at different epoch and the final performance after 10 epochs is usually significantly lower than the peak perfromance. 

Tables \ref{tab:qwen_05_grid}, \ref{tab:qwen_15_grid}, \ref{tab:deepseek_13_grid}, \ref{tab:deepseek_67_grid} and \ref{tab:codellama_grid} contains the comprehensive, epoch-by-epoch Pass@1 on MBPP across all tested model configurations.
\renewcommand{\arraystretch}{1.15}
\setlength{\tabcolsep}{10pt}

\begin{table*}[t]
\centering
\small
\begin{tabular}{lcccccccccc}
\toprule
Learning Rate & E1 & E2 & E3 & E4 & E5 & E6 & E7 & E8 & E9 & E10 \\
\midrule
$2\times10^{-6}$ & 0 & 0 & 0 & 0 & 0 & 0 & 0 & 0 & 0 & 0 \\
$1\times10^{-5}$ & 0 & 0 & 0 & 7 & 17 & 26 & 19 & 33 & 23 & 23 \\
$2\times10^{-5}$ & 0 & 1 & 17 & 44 & 46 & 40 & 38 & 2 & 16 & 2 \\
$3\times10^{-5}$ & 0 & 17 & 46 & 48 & 38 & 9 & 0 & 16 & 0 & 2 \\
$5\times10^{-5}$ & 0 & 44 & 35 & 4 & 4 & 7 & 5 & 10 & 0 & 1 \\
\bottomrule
\end{tabular}
\caption{Pass@1 (\%) chronological evaluation across training epochs for Qwen 0.5B (Batch Size 6).}
\label{tab:qwen_05_grid}
\end{table*}

\begin{table*}[t]
\centering
\small
\begin{tabular}{lcccccccccc}
\toprule
Learning Rate & E1 & E2 & E3 & E4 & E5 & E6 & E7 & E8 & E9 & E10 \\
\midrule
$1\times10^{-6}$ & 12 & 13 & 12 & 13 & 13 & 14 & 11 & 13 & 11 & 10 \\
$2\times10^{-6}$ & 14 & 13 & 13 & 13 & 11 & 9 & 10 & 9 & 9 & 9 \\
$1\times10^{-5}$ & 13 & 13 & 9 & 48 & 30 & 11 & 1 & 1 & 8 & 21 \\
$2\times10^{-5}$ & 13 & 11 & 30 & 0 & 0 & 0 & 0 & 0 & 0 & 0 \\
$3\times10^{-5}$ & 14 & 54 & 26 & 14 & 0 & 0 & 0 & 0 & 0 & 0 \\
$5\times10^{-5}$ & 10 & 39 & 32 & 0 & 1 & 0 & 0 & 0 & 0 & 0 \\
\bottomrule
\end{tabular}
\caption{Pass@1 (\%) chronological evaluation across training epochs for Qwen 1.5B (Batch Size 6).}
\label{tab:qwen_15_grid}
\end{table*}

\begin{table*}[t]
\centering
\small
\begin{tabular}{lcccccccccc}
\toprule
Learning Rate & E1 & E2 & E3 & E4 & E5 & E6 & E7 & E8 & E9 & E10 \\
\midrule
$5\times10^{-7}$ & 56 & 57 & 56 & 56 & 56 & 54 & 56 & 56 & 56 & 56 \\
$1\times10^{-6}$ & 55 & 56 & 55 & 56 & 56 & 55 & 56 & 56 & 56 & 56 \\
$2\times10^{-6}$ & 56 & 56 & 55 & 56 & 57 & 57 & 56 & 57 & 57 & 57 \\
$1\times10^{-5}$ & 55 & 55 & 57 & 56 & 52 & 52 & 49 & 50 & 49 & 47 \\
$2\times10^{-5}$ & 57 & 57 & 55 & 40 & 45 & 48 & 35 & 22 & 32 & 32 \\
$3\times10^{-5}$ & 56 & 56 & 39 & 46 & 44 & 29 & 10 & 23 & 15 & 25 \\
$5\times10^{-5}$ & 58 & 46 & 22 & 12 & 32 & 20 & 0 & 3 & 15 & 17 \\
\bottomrule
\end{tabular}
\caption{Pass@1 (\%) chronological evaluation across training epochs for DeepSeek 1.3B (Batch Size 6).}
\label{tab:deepseek_13_grid}
\end{table*}

\begin{table*}[t]
\centering
\small
\begin{tabular}{lcccccccccc}
\toprule
Learning Rate & E1 & E2 & E3 & E4 & E5 & E6 & E7 & E8 & E9 & E10 \\
\midrule
$5\times10^{-7}$ & 4 & 2 & 3 & 2 & 2 & 3 & 3 & 1 & 2 & 2 \\
$1\times10^{-6}$ & 3 & 2 & 3 & 1 & 3 & 2 & 2 & 3 & 3 & 3 \\
$2\times10^{-6}$ & 3 & 1 & 2 & 2 & 2 & 2 & 3 & 2 & 2 & 1 \\
$1\times10^{-5}$ & 2 & 2 & 9 & 16 & 14 & 15 & 9 & 14 & 11 & 11 \\
$2\times10^{-5}$ & 0 & 3 & 29 & 34 & 17 & 6 & 5 & 11 & 11 & 11 \\
$3\times10^{-5}$ & 2 & 26 & 0 & 0 & 0 & 0 & 0 & 0 & 0 & 0 \\
$5\times10^{-5}$ & 10 & 0 & 0 & 0 & 0 & 0 & 0 & 0 & 0 & 0 \\
\bottomrule
\end{tabular}
\caption{Pass@1 (\%) chronological evaluation across training epochs for DeepSeek 6.7B (Batch Size 2).}
\label{tab:deepseek_67_grid}
\end{table*}

\begin{table*}[t]
\centering
\small
\begin{tabular}{lcccccccccc}
\toprule
Learning Rate & E1 & E2 & E3 & E4 & E5 & E6 & E7 & E8 & E9 & E10 \\
\midrule
$2\times10^{-6}$ & 0 & 0 & 0 & 0 & 1 & 0 & 0 & 1 & 0 & 0 \\
$1\times10^{-5}$ & 0 & 7 & 1 & 8 & 0 & 1 & 1 & 0 & 0 & 0 \\
$2\times10^{-5}$ & 1 & 0 & 0 & 0 & 0 & 0 & 0 & 0 & 0 & 0 \\
$3\times10^{-5}$ & 9 & 0 & 0 & 0 & 0 & 0 & 0 & 0 & 0 & 0 \\
$5\times10^{-5}$ & 8 & 0 & 0 & 0 & 0 & 0 & 0 & 0 & 0 & 0 \\
\bottomrule
\end{tabular}
\caption{Pass@1 (\%) chronological evaluation across training epochs for CodeLlama 7B (Batch Size 2).}
\label{tab:codellama_grid}
\end{table*}

\section{Logit Gap} \label{sec:logit_gap}
\begin{figure}[h]
\centering
\includegraphics[width=\columnwidth]{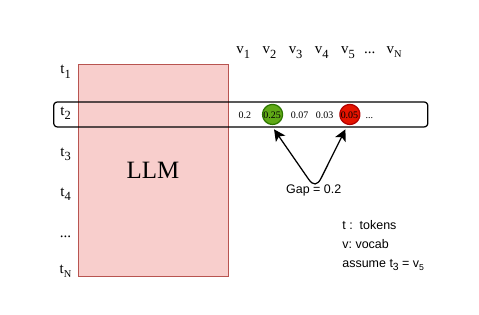}
\caption{Difference between predicted token and offline token. The output of LLM is not normalized and so the gap is unbounded.}
\label{fig:logit gap}
\end{figure}

Figure \ref{fig:logit gap} provides a visual illustration of logit gap between model preferred token and dataset token. 

\begin{figure*}[h!]
\centering
    \begin{subfigure}[b]{0.99\textwidth}
     \includegraphics[width=0.33\linewidth]{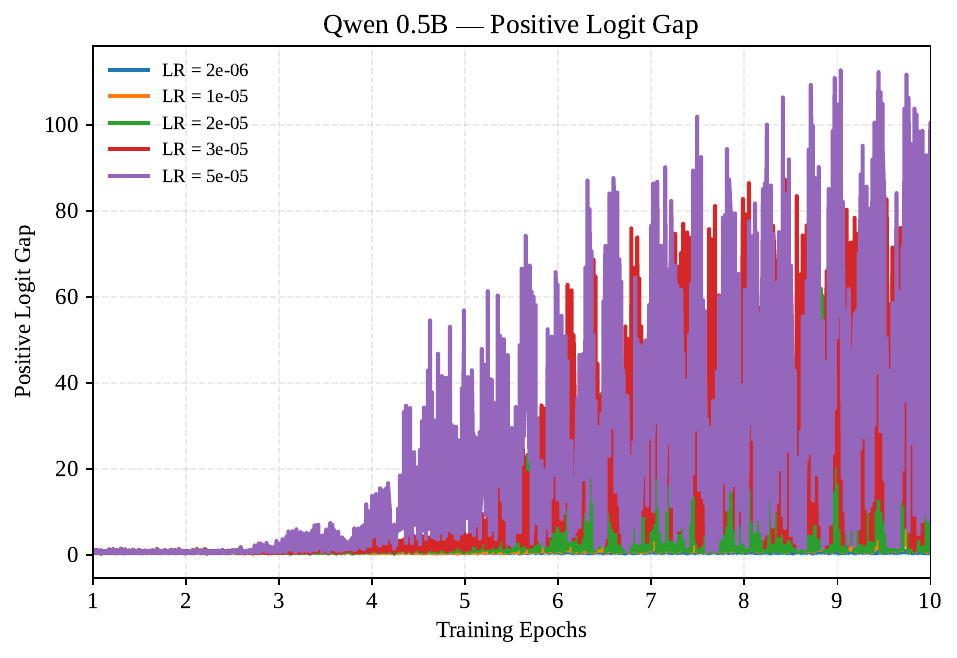} 
     \includegraphics[width=0.33\linewidth]{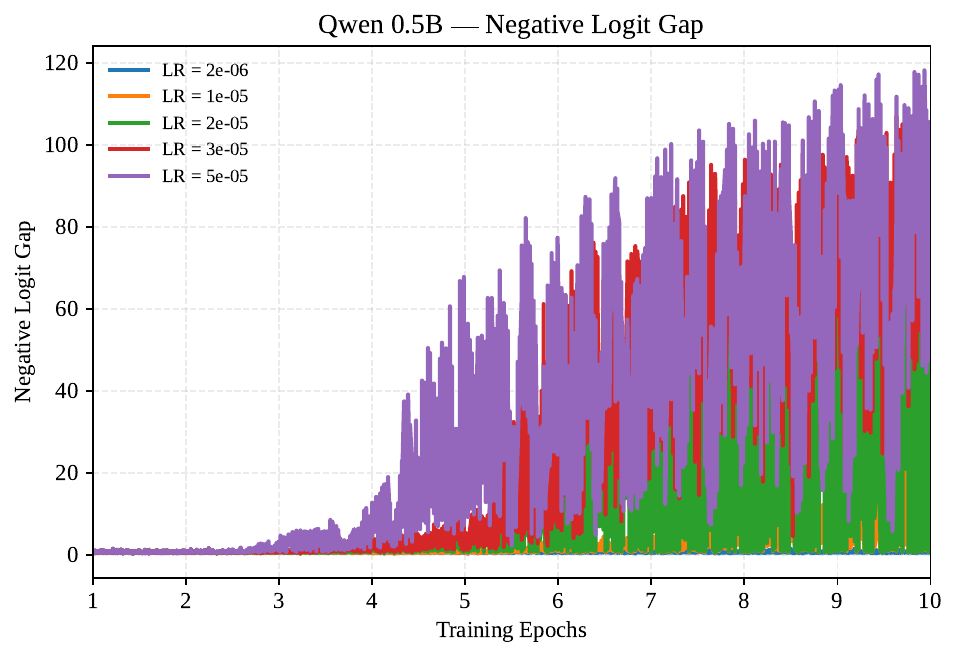}
     \includegraphics[width=0.33\linewidth]{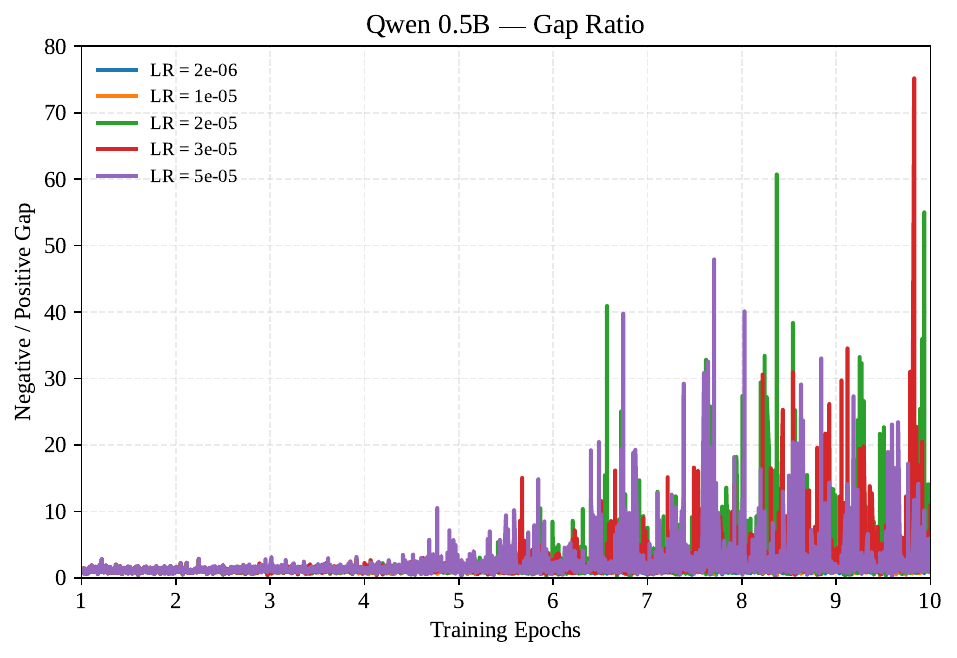}
    \end{subfigure}

    \begin{subfigure}[b]{0.99\textwidth}
     \includegraphics[width=0.33\linewidth]{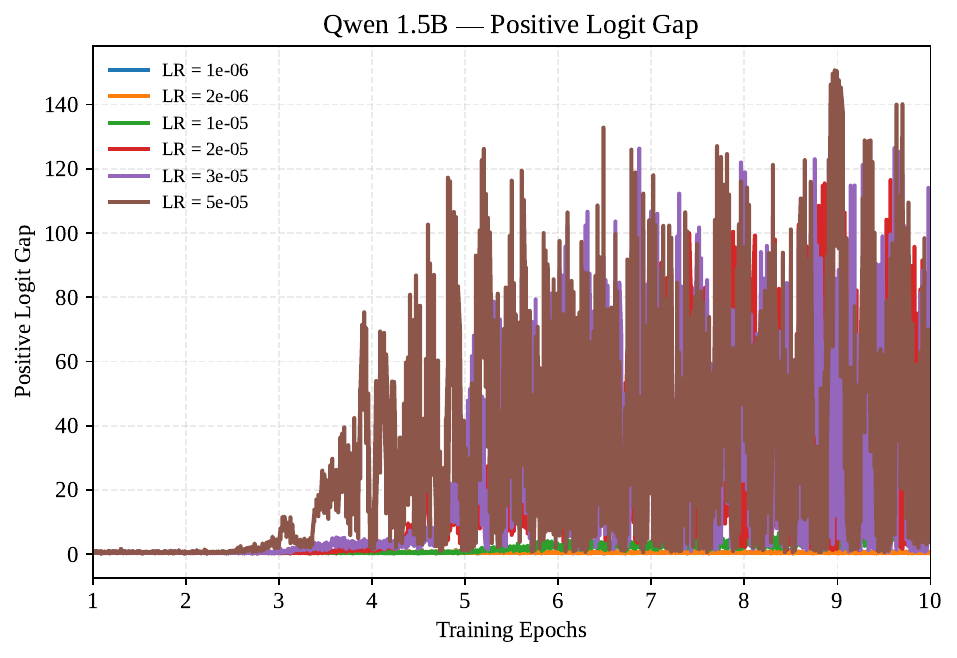} 
     \includegraphics[width=0.33\linewidth]{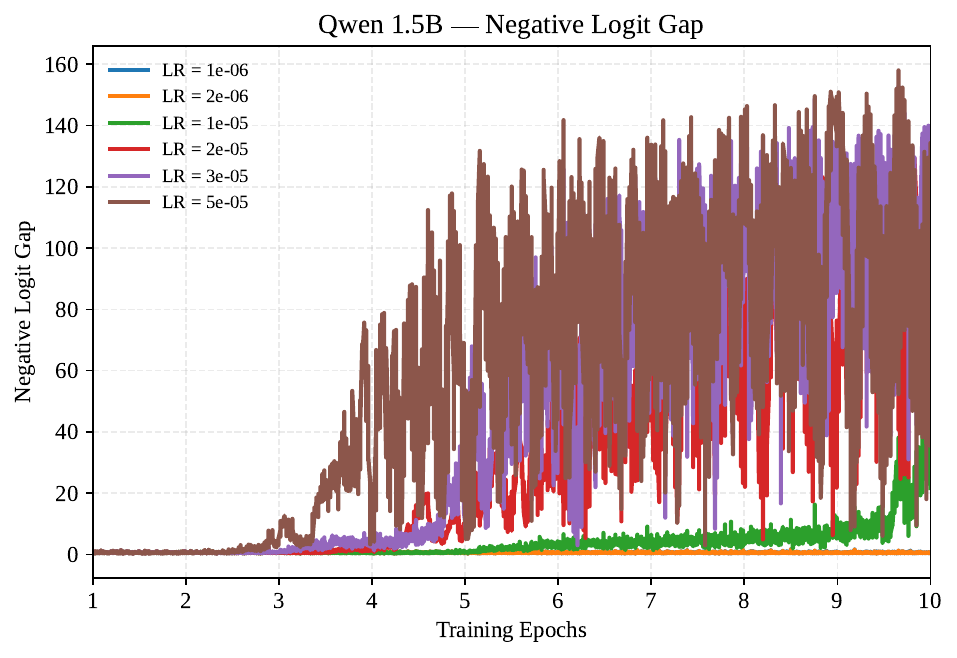}
     \includegraphics[width=0.33\linewidth]{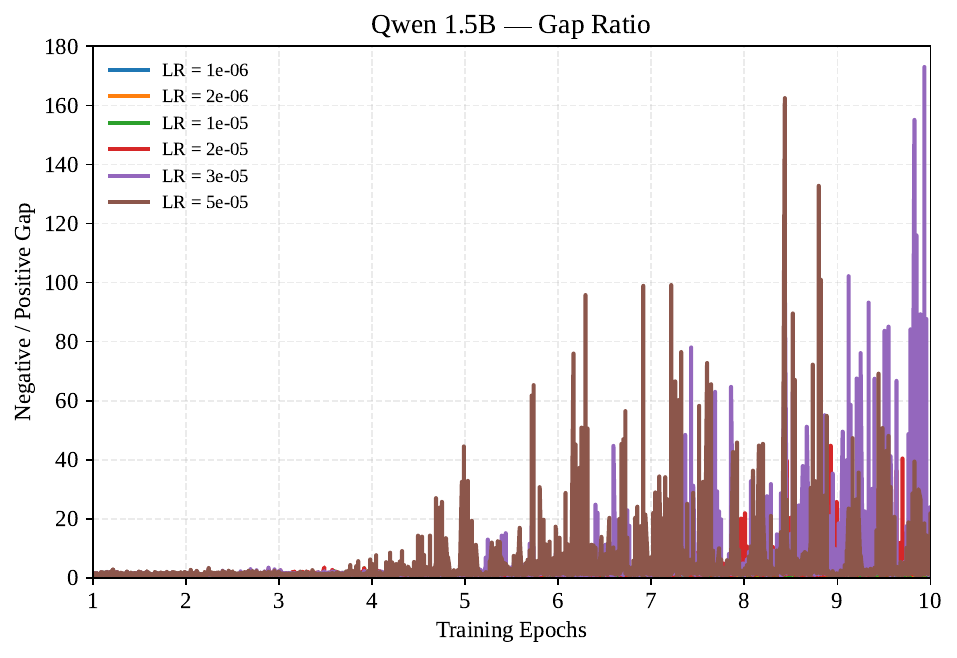}
    \end{subfigure}

    \begin{subfigure}[t]{0.99\textwidth}
    \includegraphics[width=0.33\linewidth]{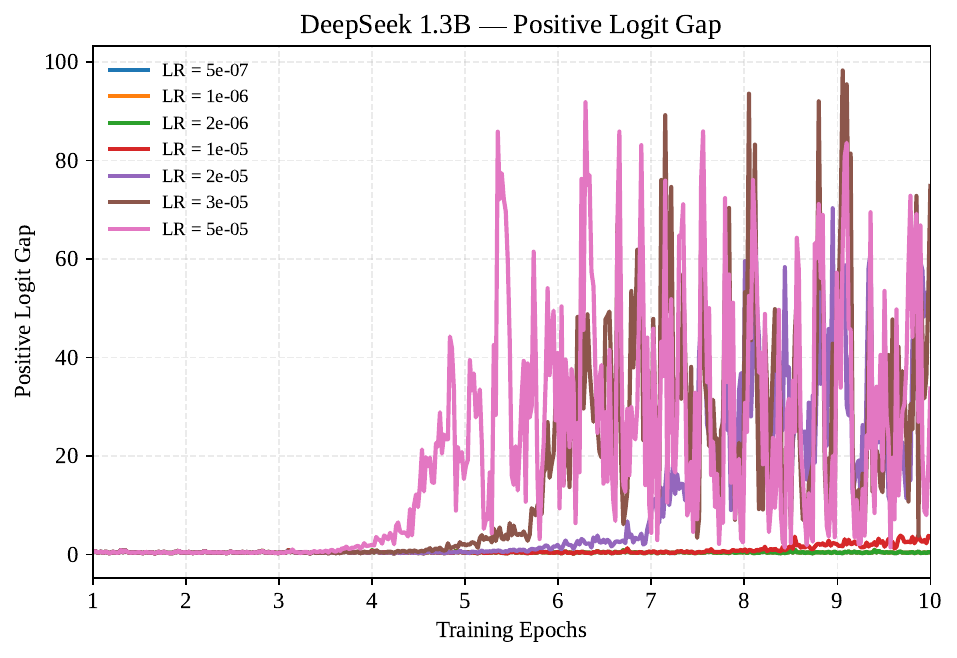}
    \includegraphics[width=0.33\linewidth]{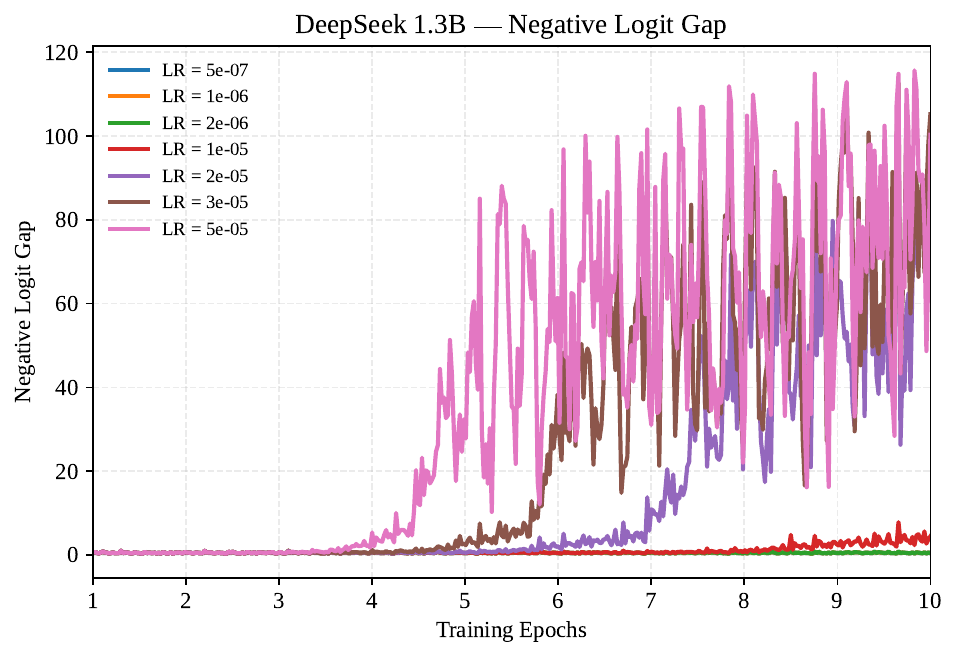}
    \includegraphics[width=0.33\linewidth]{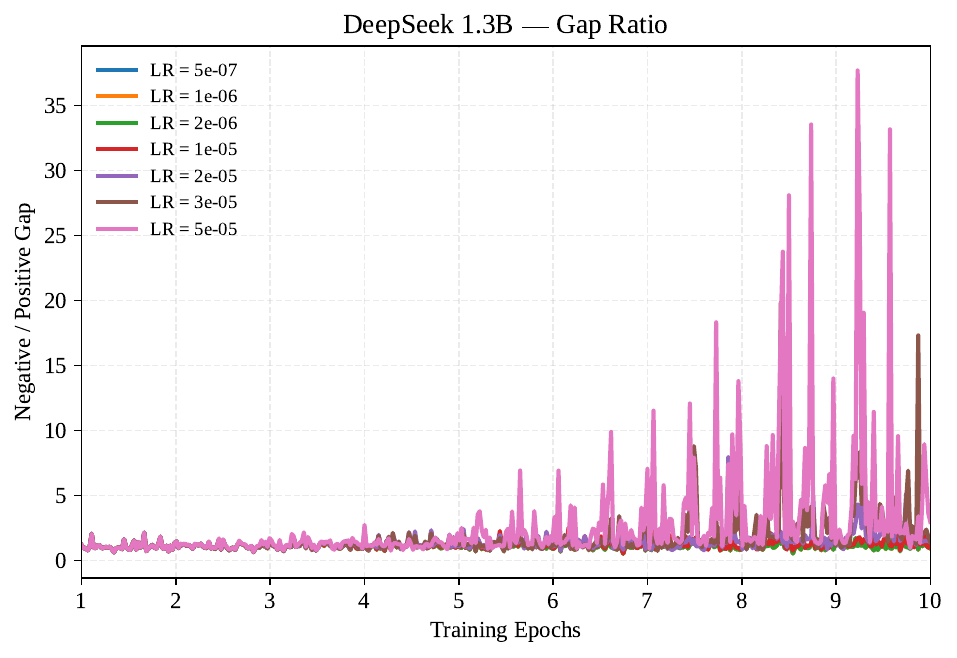}
    \end{subfigure}
    \begin{subfigure}[t]{0.99\textwidth}
    \includegraphics[width=0.33\linewidth]{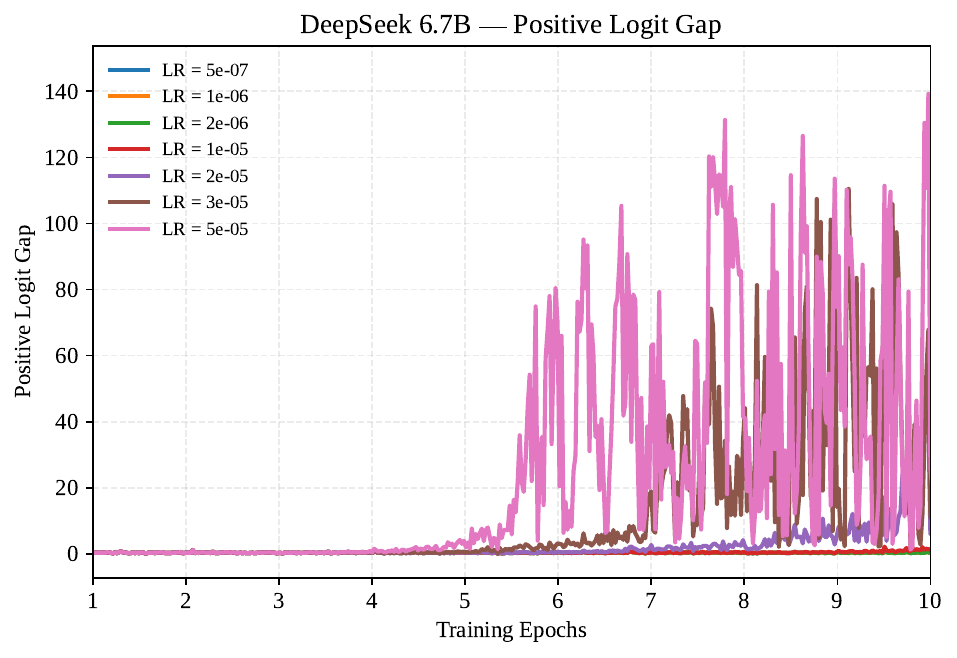}
    \includegraphics[width=0.33\linewidth]{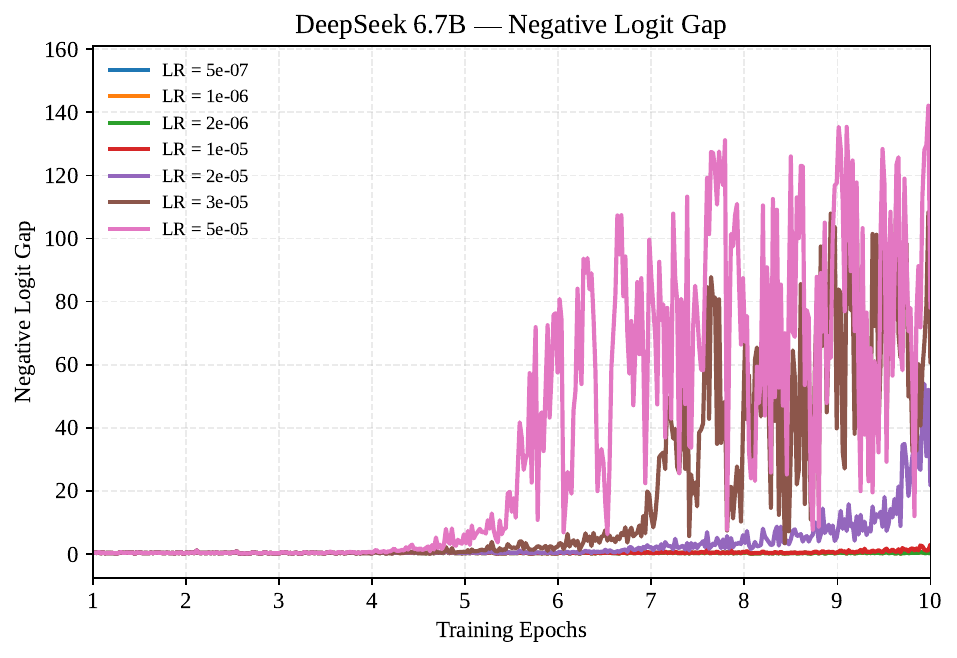}
    \includegraphics[width=0.33\linewidth]{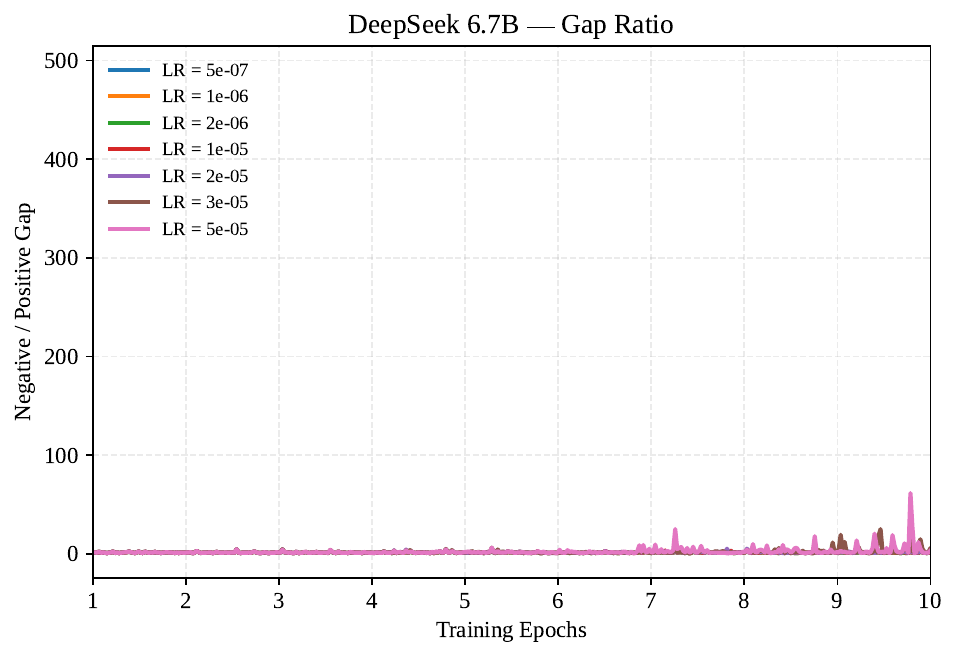}
    \end{subfigure}
    \begin{subfigure}[b]{0.99\textwidth}
     \includegraphics[width=0.33\linewidth]{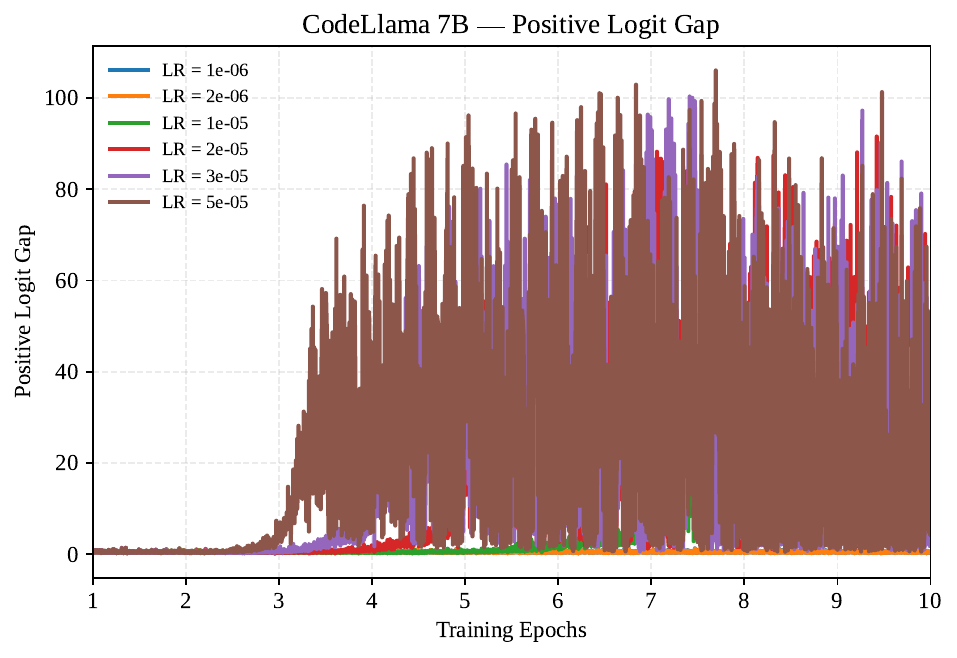} 
     \includegraphics[width=0.33\linewidth]{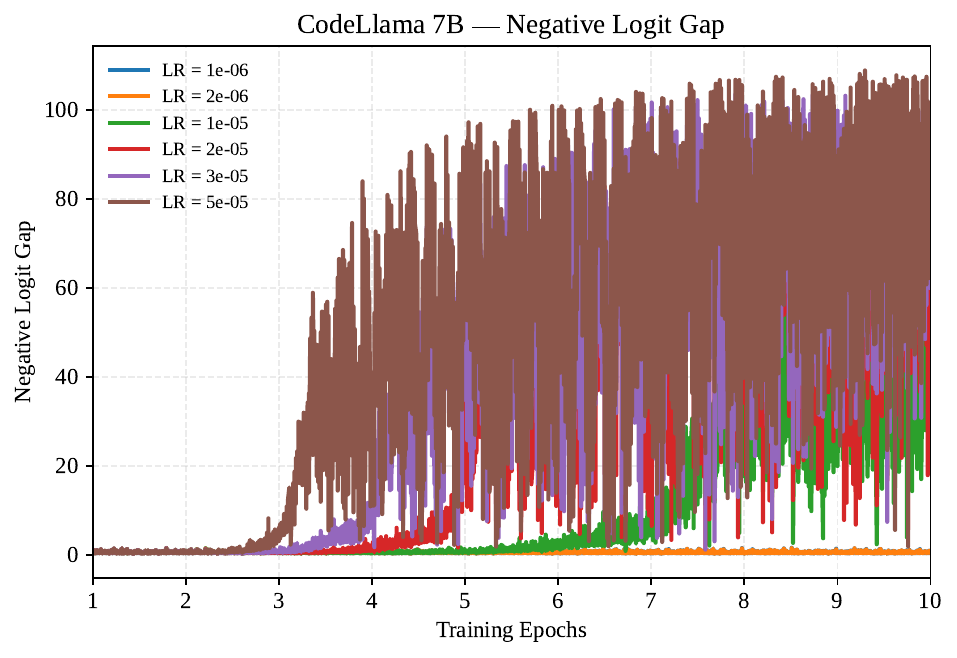}
     \includegraphics[width=0.33\linewidth]{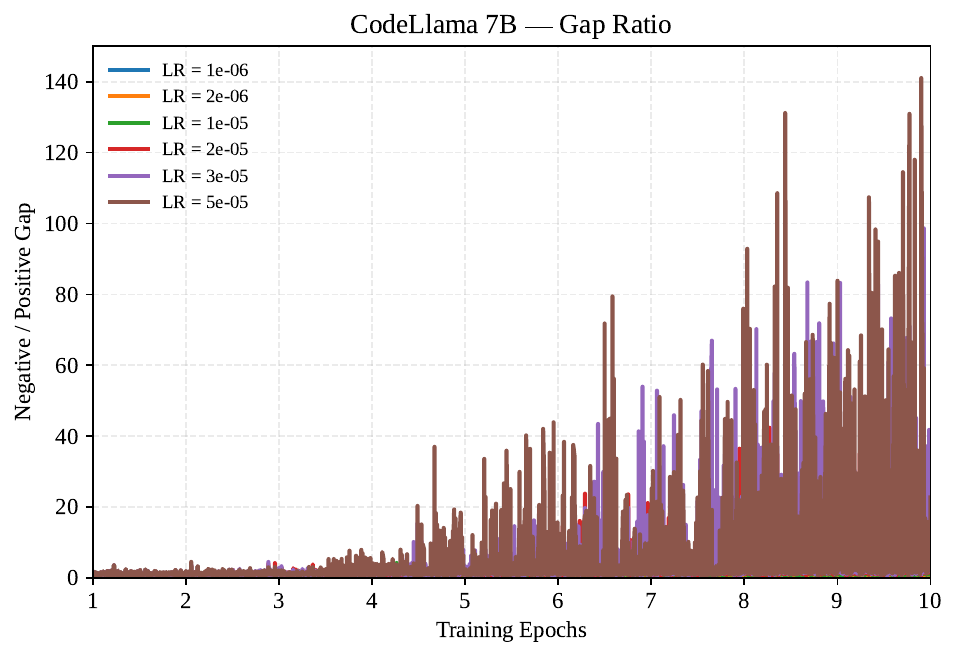}
    \end{subfigure}
    
\caption{Negative gap, positive gap and gap ration during training of various model.}
\label{fig:logit_gap_all}
\end{figure*}

Figure \ref{fig:logit_gap_all} shows the positive and negative gap and the gap ratio for the models other than Qwen 7B. While the plot for every model varies, they follow similar pattern. The initial positive and negative gaps are close to zero for all models, which satisfies our hypothesis in Section \ref{hypothesis}. Negative gap more than the positive gap for all models except the DeepSeek 6.7B model. Further, if the negative and positive gap remains close to zero throughout, the performance of the model does not improve. Increase in negative gap or gap ratio corresponds to degradation in performance.

\section{Case Studies}
\subsection{Example 1 (Qwen 0.5B)}

\textbf{Problem:}
\emph{Write a function that checks whether a string represents a valid integer.}

\medskip 

\textbf{Base Model Output (Failed)}
\begin{verbatim}
Write a function to check if a string 
represents an integer or not.
Your code should pass the following test:
assert check_integer("123")==True
indexPath
(indexPath
...

\end{verbatim}
\medskip 

\textbf{Offline RL Model Output}

\begin{verbatim}
def check_integer(s):
    try:
        int(s)
        return True
    except ValueError:
        return False
\end{verbatim}

\subsection{Example 2 (Qwen 0.5B)}

\textbf{Problem:}
\emph{Write a function that returns the frequency of each element in a list as a dictionary.}

\medskip

\textbf{Base Model Output (Failed)}
\begin{verbatim}
Write a function to get the frequency of all 
the elements in a list, returned as a 
dictionary. Your code should pass the 
following test:
assert freq_count([10,10,10,10,20,20,20,
20,40,40,50,50,30])==
({10:4,20:4,40:2,50:2,30:1})
orical
countertops
...
\end{verbatim}

\medskip

\textbf{Offline RL Model Output}

\begin{verbatim}
def freq_count(lst):
    freq = {}
    for num in lst:
        if num in freq:
            freq[num] += 1
        else:
            freq[num] = 1
    return freq
\end{verbatim}

\textbf{Discussion}

The base model produces incomplete and non-executable output, failing to generate a valid Python solution. After offline RL training, the model generates a correct implementation.

\section{Effect of Different Data Sampling} \label{resampling}

To evaluate the effect of different data sampling, we repeated the post-training experiments using alternative samples from the offline dataset. For each replacement dataset, we kept the same set of programming problems but replaced the code submissions with different solutions sampled from CodeNet while maintaining a similar status distribution. These experiments were conducted using the Qwen-0.5B model with a learning rate of $3\times10^{-5}$, while keeping all other training hyperparameters unchanged.

In this set of experiments, the best checkpoint obtained from training on the original dataset achieved an MBPP Pass@1 of 42. This checkpoint was then used to initialize training on the first replacement dataset. The best checkpoint from this experiment was subsequently used to initialize training on the next replacement dataset. The performance of Qwen 0.5B improves from 0 for the base model to 42 with first offline training and then to 46 with new samples for the same problems.   

\begin{figure*}[t]
    \centering

    \begin{subfigure}[t]{0.32\textwidth}
        \centering
        \includegraphics[width=\linewidth]{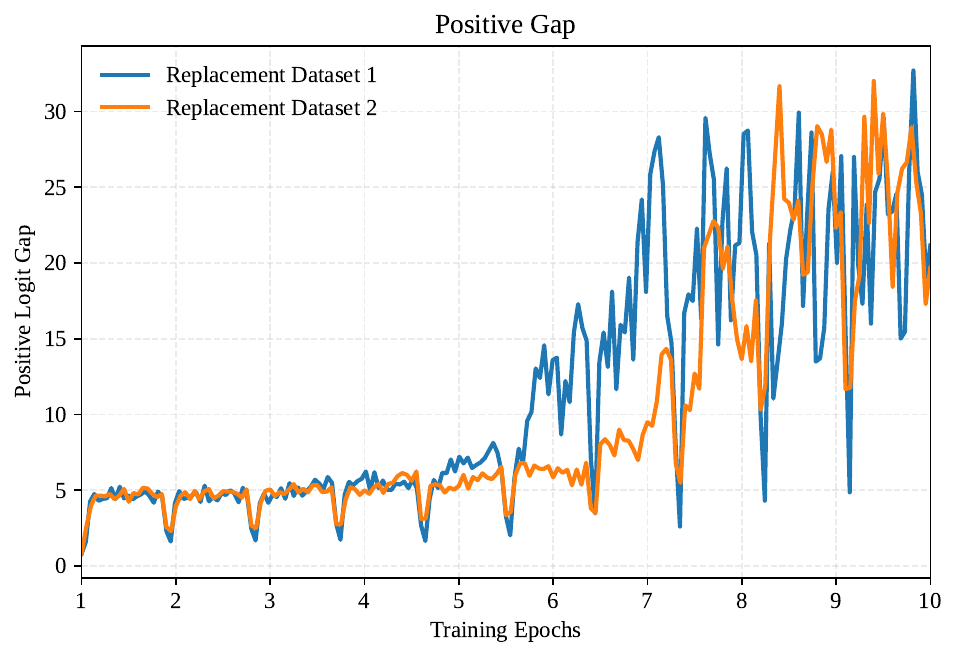}
        \caption{Positive Logit Gap}
        \label{fig:positive_gap}
    \end{subfigure}
    \hfill
    \begin{subfigure}[t]{0.32\textwidth}
        \centering
        \includegraphics[width=\linewidth]{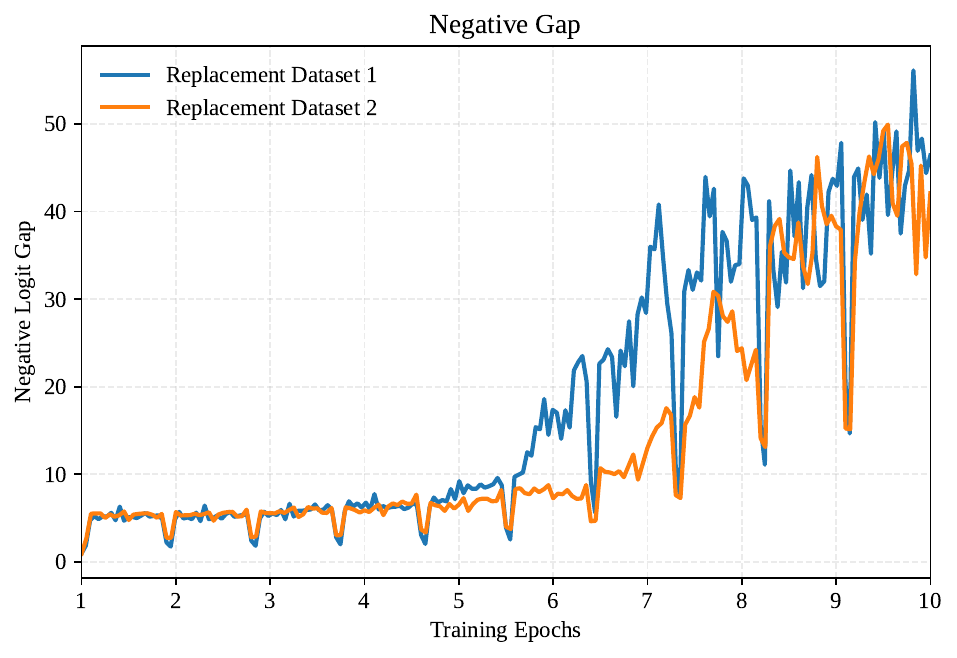}
        \caption{Negative Logit Gap}
        \label{fig:negative_gap}
    \end{subfigure}
    \hfill
    \begin{subfigure}[t]{0.32\textwidth}
        \centering
        \includegraphics[width=\linewidth]{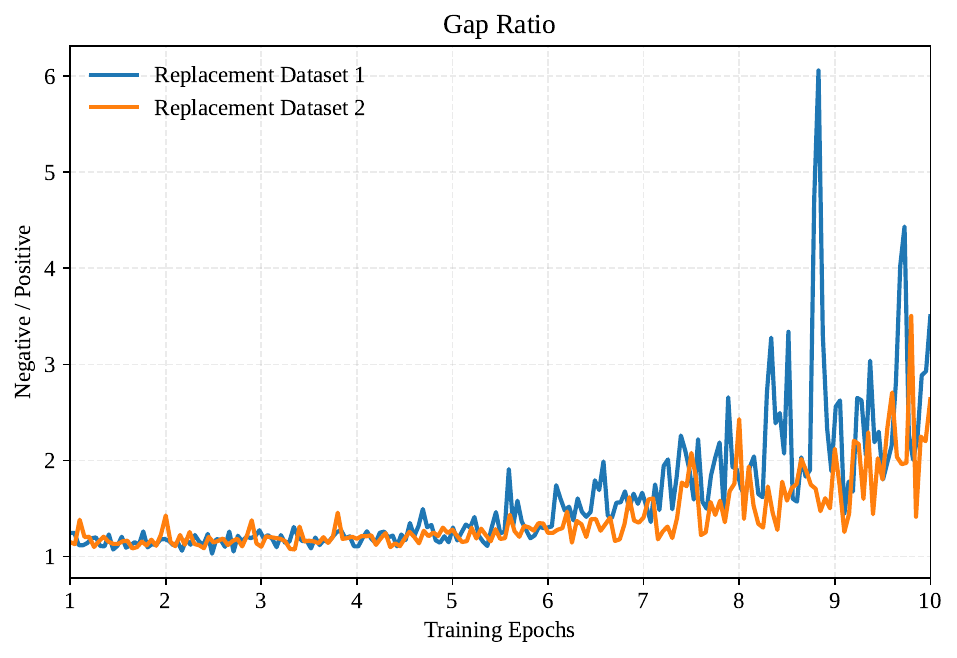}
        \caption{Gap Ratio}
        \label{fig:gap_ratio}
    \end{subfigure}

    \caption{Evolution of the positive logit gap, negative logit gap, and gap ratio during post-training on two replacement datasets. Similar trends are observed across different data samplings.}
    \label{fig:replacement_gap}
\end{figure*}

As shown in Table~\ref{tab:replacement_mbpp}, post-training on the replacement datasets led to a further increase in MBPP Pass@1. We also observed similar logit-gap trends across the different data samplings (Figure~\ref{fig:replacement_gap}), suggesting that the training behavior remains consistent when the offline training samples are changed. Importantly, the logit gap after taking in new samples is again close to zero, suggesting that considering new samples again makes the training effectively on-policy.

\begin{table*}[t]
\centering
\small
\begin{tabular}{lcccccccccc}
\toprule
\textbf Dataset & E1 & E2 & E3 & E4 & E5 & E6 & E7 & E8 & E9 & E10 \\
\midrule
Replacement Dataset 1 & 46 & 43 & 41 & 43 & 45 & 39 & 17 & 13 & 8 & 1 \\
Replacement Dataset 2 & 45 & 43 & 44 & 42 & 43 & 42 & 34 & 26 & 34 & 31 \\
\bottomrule
\end{tabular}
\caption{MBPP Pass@1 across training epochs for post-training on replacement datasets using the Qwen-0.5B model ($\mathrm{LR}=3\times10^{-5}$).}
\label{tab:replacement_mbpp}
\end{table*}

\end{document}